\documentclass[journal,10pt]{IEEEtran}
\usepackage{amsthm}
\usepackage{amsmath}
\usepackage{graphicx}
\usepackage{algorithm}
\usepackage{algorithmicx}
\usepackage{algpseudocode}
\usepackage{subfigure}
\usepackage{overpic}
\usepackage{colortbl}
\ifCLASSINFOpdf
\else
\fi
\begin{document}
\title{A Residual Solver and Its Unfolding Neural Network for Total Variation Regularized Models}

\author{Yuanhao~Gong% <-this % stops a space
	\thanks{Manuscript received April 19, 2005; revised September 17, 2014.}
	}


\markboth{Journal of \LaTeX\ Class Files,~Vol.~14, No.~8, August~2015}%
{Yuanhao: RSnet}



\maketitle

\begin{abstract}
This paper proposes to solve the Total Variation regularized models by finding the residual between the input and the unknown optimal solution. After analyzing a previous method, we developed a new iterative algorithm, named as Residual Solver, which implicitly solves the model in gradient domain. We theoretically prove the uniqueness of the gradient field in our algorithm. We further numerically confirm that the residual solver can reach the same global optimal solutions as the classical method on 500 natural images. Moreover, we unfold our iterative algorithm into a convolution neural network (named as Residual Solver Network). This network is unsupervised and can be considered as an ``enhanced version'' of our iterative algorithm. Finally, both the proposed algorithm and neural network are successfully applied on several problems to demonstrate their effectiveness and efficiency, including image smoothing, denoising, and biomedical image reconstruction. The proposed network is general and can be applied to solve other total variation regularized models.
\end{abstract}


% \fi
\IEEEpeerreviewmaketitle

\section{Introduction}
%%%%%%%%% BODY TEXT
Various image processing and computer vision tasks are ill-posed, such as image denoising, biomedical image reconstruction, image smoothing, segmentation, object recognition, etc. The illness comes from the fact that it is impossible to recover the ground truth without making any assumption about it. For example, in the image denoising task, it is impossible to tell if a pixel intensity contains noise or not without making any assumption about the ground truth. Another example is the image smoothing. It is not clear what should be smoothed out without any prior knowledge. In the biomedical image reconstruction, the input signal and the desired output even do not work in the same space. And the illness comes from the fact that there are not enough observations available due to physical limitations, such as dose restriction, equipment angle limitation, temperature requirement, pressure condition, etc.

Although ill-posed problems are common, solving these problems is challenging. Usually, there are many solutions that satisfy the constraints in the ill-posed problems. In mathematical words, the solution space is too large to find the desired one solution. Moreover, the computational cost for such problems is usually very high. Therefore, it is not realistic to search the full solution space in practice.

To solve such ill-posed problems, a prior or an assumption about the ground truth must be imposed, reducing the solution space or even making the solution unique. For example, the well-known total variation regularization leads to a piecewise constant signal~\cite{TV1992}. The ground truth signal might be assumed to be piecewise linear, leading to the famous guided filter~\cite{he2010guided}. The ground truth might be assumed to be a minimal surface, leading to the area regularization~\cite{gong:Bernstein,gong:cf,GONG2019329}.  The ground truth might be assumed to be piecewise developable, leading to the Gaussian curvature regularization~\cite{gong2013a,gong2009symmetry}. Different assumptions usually lead to different resulting images, which visually reflects the imposed priors. For example, the well-known total variation regularization will make the resulting image piecewise constant, which is known as staircase artifacts. These artifacts reflect the imposed prior. Such prior is usually imposed by a regularization term, which is a mathematical model for the imposed prior or assumption. 

The same prior or assumption, however, might have several different regularization forms because it can be modeled by different mathematical formula with various accuracy and computation complexity. For example, the well-known gradient distribution prior can be modeled by Gaussian distribution, Hyper Laplace distribution~\cite{krishnan2009fast} or the derivative of $atan$ function~\cite{Gong:2014a,gong:gdp}, etc. These models impose the same gradient distribution prior, but they have different approximation accuracy and different optimization properties, such as convexity, fast computation, explicit gradient formula, etc.

\subsection{Total Variation Regularization}
Among all regularization terms, Total Variation (TV) might be the most popular one because of its simplicity and effectiveness~\cite{TV1992}. Assuming the unknown ground truth signal is $U(\vec{x})$ where $\vec{x}$ is the spatial coordinate, the traditional total variation regularization term is
\begin{equation}
TV(U(\vec{x}))= \|\nabla U(\vec{x})\|_p\,,
\end{equation} where $\nabla$ is the gradient operator and $p=\{1,2\}$ indicates the standard $\ell_p$ norm. This term is isotropic when $p=2$ and becomes anisotropic when $p=1$. Choosing which $\ell_p$ norm is problem specific. In the rest of this paper, we use $p=1$ because it has been shown preserving edges better in many practical applications~\cite{gong:gdp}.

Although there are some variants of TV, such as spatial weighted TV and $\ell_{0.5}$ norm TV (Hyper Laplace), the standard TV with $\ell_{1}$ norm still is the most popular regularization term. The reasons come from three aspects. First, from optimization point of view, this term is convex. This means the global optimal solution is unique. And the convex optimization algorithms can be used for this term. Second, the $\ell_{1}$ norm is a simple computation operation. Thus, it can be efficiently computed and evaluated in practice. Third, the anisotropy property of this term is important in preserving edges in image processing. In general, preserving edges is always preferred in many image processing tasks.

When TV is used as the regularization term, many image processing problems, such as denoising, segmentation, inpainting and fusion, etc, can be formulated as the following minimization problem: 
\begin{equation}
\label{eq:loss}
\min_{U}\left\{\frac{1}{2}\|AU(\vec{x})-f(\vec{x})\|_2^2+\lambda TV(U(\vec{x}))\right\}
\end{equation} where $A$ is an imaging matrix, $U$ is the unknown signal to be estimated, $\vec{x}$ is a spatial coordinate, $f$ is the observed signal and $\lambda>0$ is a scalar that balances the imaging and the TV prior ($\lambda$ usually is related with noise level). The imaging matrix $A$ models a linear imaging process, such as blurring, down sampling, up sampling, etc. A linear matrix already covers a large range of image processing problems.  

\subsection{Iterative Solvers}
Many solvers have been developed for this model (Equation~\ref{eq:loss}). The earliest iterative scheme was proposed in ~\cite{TV1992}. This method is slow due to the stability constraints in the time step size. To accelerate the optimization procedure, a number of alternative methods have been proposed. One popular method is primal dual solver~\cite{Chan:1999,chambolle:2011}. Primal dual method simultaneously finds the optimal solution for this model and its dual, improving the numerical stability. Another technique is to split the imaging model and the TV regularization, by introducing a variable $W$:
\begin{equation}
\min_{U}\left\{\frac{1}{2}\|AU-f\|_2^2+\lambda TV(W) \right\}~\mathrm{s.t.}~W=U\,.
\end{equation}
This new model can be solved by alternating direction method of multipliers (ADMM)~\cite{Wahlberg:2012}, split Bregman~\cite{goldstein2009split}, etc.  Equation~\ref{eq:loss} can also be solved by Iterative Shrinkage and Thresholding Algorithm (ISTA) ~\cite{Beck2009}. These methods are based on nonlinear partial differential equations and thus time consuming, compared with recently developed neural network solvers.

\subsection{Neural Network Solvers}
\label{sec:nn}
With the development of deep learning, several neural networks have been proposed to solve Eq.~\ref{eq:loss}. One of the earliest work is the Learned ISTA (LISTA), which unfolds the ISTA as neural networks to perform the optimization algorithm~\cite{gregor2010learning}. One layer in the network corresponds to one iteration in ISTA. Different from ISTA, all the parameters in LISTA implicitly communicate with each other during the training process. Such communication can be considered as high order scheme from the partial differential equation point of view. Therefore, the network can solve the problem within a small number of layers, in contrast with the large number of iterations in iterative algorithm ISTA.

ISTA-net is another extension of ISTA~\cite{ISTA}. It unfolds the ISTA but keeps the symmetric network structure, requiring the encoding and decoding are inverse functions for each other. Such constraint makes the network invertable, leading to better numerical stability than LISTA.

Another unfolding method is the primal dual network, which unfolds the primal dual iteration for Eq.~\ref{eq:loss}~\cite{Adler2018}. It has been shown converging much faster than the original primal dual iteration. 

ADMM-net unfolds the classical ADMM iteration, leading to a neural network representation~\cite{sun2016deep}. It first trains one layer. Then it fixes this layer, adds another layer and only trains the new added layer. By repeating this procedure, it shows good performance for MRI reconstruction problem. In this network, after 10 layers, adding more layers does not improve the result anymore.

Recently, a CNN based image reconstruction method for inverse problem is proposed in~\cite{jin2017deep}. It shows that neural networks are very efficient in solving image reconstruction problems thanks to the parameter communication between different layers. 

In general, neural network methods are more efficient than the iterative algorithms because of two reasons. First, the parameters in the neural networks can communicate with each other during the training process. This corresponds to the high order scheme from partial differential equation point of view. Second, the parameters in the neural networks are more adaptive to the training data. In contrast, the operations in the iterative algorithms are independent from the input image. As a result, the neural network usually only needs a small number of layers to solve Eq.~\ref{eq:loss} (computational efficiency).

\subsection{Motivations and  Contributions}
Most of these neural network algorithms are supervised, requiring the ground truth to be explicitly given. Such requirement is not easily satisfied in most practical applications. This motivates us to develop an unsupervised neural network. Our contributions fall in following two aspects:  
\begin{itemize}
	\item We develop a new iterative algorithm that solves the Eq.~\ref{eq:loss}. Instead of directly finding the optimal solution, we find its related residual (details will be explained in later sections). Therefore, our solver is called Residual Solver (RS).
	
	\item We unfold the proposed iterative residual solver into a neural network and our network is unsupervised. Therefore, it can be trained without knowing the optimal solution. Another advantage of our network is that it can be trained on low resolution images but applied on high resolution images. Our experiments confirm this property.
\end{itemize}

\section{Mathematical Background}
Equation~\ref{eq:loss} can be solved by many algorithms as mentioned in previous sections. Here, we give the general procedure that solves Eq.~\ref{eq:loss} and then a typical algorithm for solving the bottleneck sub problem in this procedure.

In general, Eq.~\ref{eq:loss} can be solved by the following iteration (proximal gradient method~\cite{Boyd2011})
\begin{eqnarray}
\label{eq:origPoint}
V^{t+1}&=&U^t-\frac{1}{\alpha}A^T(AU^t-f-d^{t})\\
\label{eq:middle1}
U^{t+1}&=&\arg\min_U\left\{\frac{\alpha}{2}\|U-V^{t+1}\|_2^2+\lambda TV(U)\right\}\\
d^{t+1}&=&d^t+f-AU^{t+1}\,,
\end{eqnarray} where $\alpha>2\|A^TA\|$ is a parameter, $d$ is the scaled dual variable and $t$ is the iteration index. The initial values are $d^0=0$ and $U^0=0$.  

\subsection{Dual First Iteration}
Since this procedure is iterative, we can modify this iteration by updating the dual variable $d$ first, leading to a more efficient algorithm
\begin{eqnarray}
\label{eq:dual}
d^{t+1}&=&d^t+f-AU^{t}\\
\label{eq:point}
V^{t+1}&=&U^t+\frac{1}{\alpha}A^T(2d^{t+1}-d^t)\\
\label{eq:middle}
U^{t+1}&=&\arg\min_U\left\{\frac{\alpha}{2}\|U-V^{t+1}\|_2^2+\lambda TV(U)\right\}\,.
\end{eqnarray} In this new iteration, the term $2d^{t+1}-d^t$ in Eq.~\ref{eq:point} can be considered as extrapolation. And we also avoid the matrix multiplication with $A^TA$ in Eq.~\ref{eq:origPoint}. In fact, Eq.~\ref{eq:point} is simpler to compute, compared with Eq.~\ref{eq:origPoint}. Thus, it is faster in practice.

\subsection{The Bottleneck}
In both iteration methods, Eq.~\ref{eq:middle} is the computation bottleneck. This equation is known as the standard ROF model~\cite{TV1992}. Many algorithms have been developed to efficiently solve Eq.~\ref{eq:middle}. Among these algorithms, following iteration was proposed in~\cite{jia2010fast}
\begin{eqnarray}
\label{eq:jiab}
\vec{b}^{in+1}&=&cut(\nabla U^{in}+\vec{b}^{in},\beta)\\
\label{eq:jiaU}
U^{in+1}&=&V^{t+1}-\lambda\beta(\nabla^T\vec{b}^{in+1})\,,
\end{eqnarray} where $0<\beta<1/4$ is a scalar parameter, $in$ is an inner iteration index, $\vec{b}=(b_x,b_y)$ corresponds to the gradient field, $\vec{b}^0=0$, and $cut$ function is defined as
\begin{eqnarray}
cut(x,\beta)=
\begin{cases}
\beta, &x>\beta\cr x, &|x|\le \beta \cr -\beta, &x<-\beta\end{cases}\,.
\end{eqnarray}
This iteration can theoretically guarantee the numerical convergence and has  shown good performance in theory and practice~\cite{jia2010fast}. We refer this algorithm as FastSolver (FS) in this paper. 

\begin{figure}[!tbh]
	\includegraphics[width=0.99\linewidth]{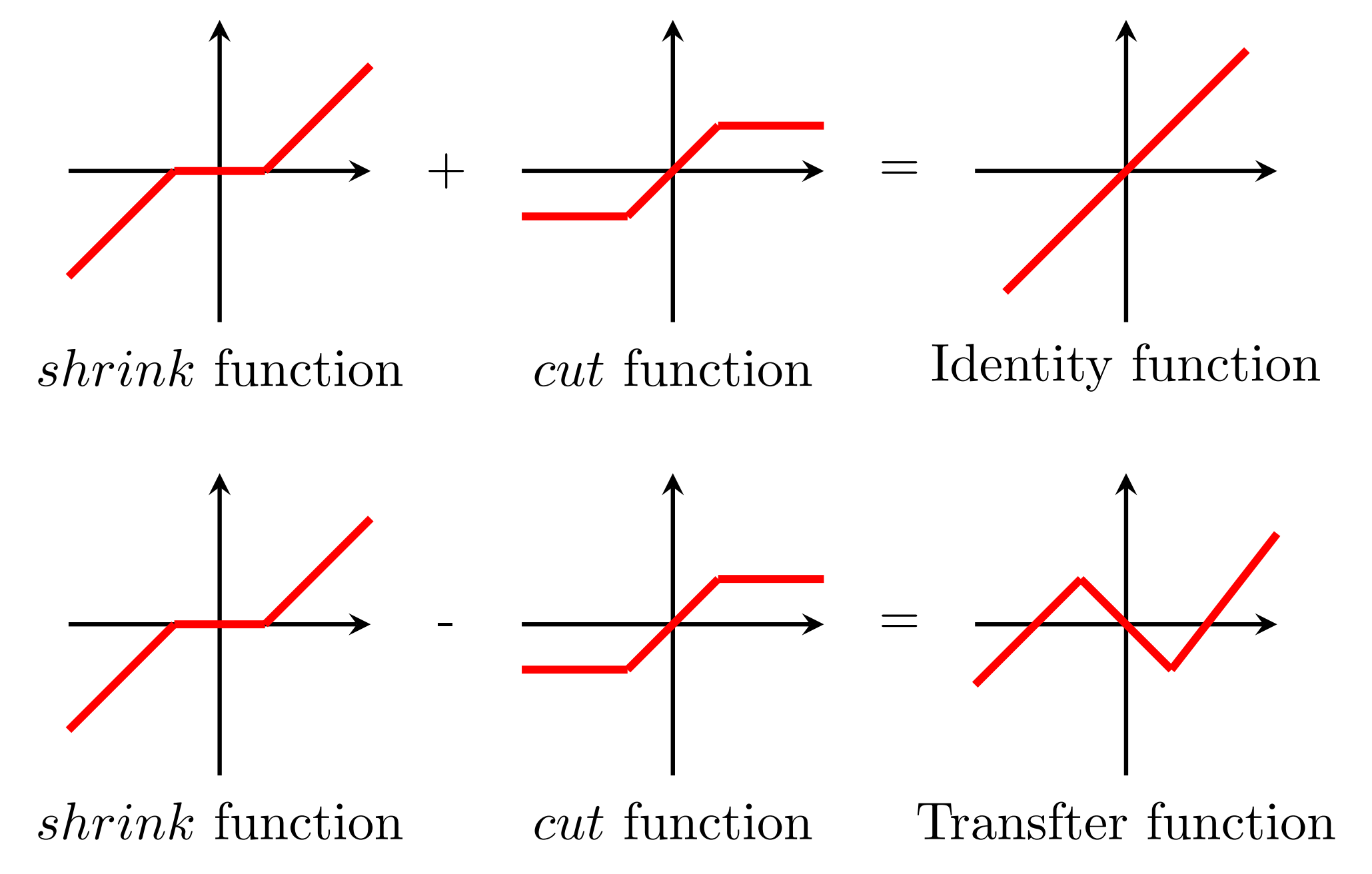}
	\caption{The relationship between $cut$ and $shrink$ functions.} \label{fig:cutshrink}
\end{figure}

This $cut$ function proposed in FS is the residual of the well-known  $shrink$ operation. As shown in Fig.~\ref{fig:cutshrink}, $cut$ function plus $shrink$ function equals the identity function. Their difference forms a nonlinear transfer function that also appears in deep learning community. Moreover, the $cut$ function is similar with several well-known activation functions from neural network field, such as sigmoid, $atan$, soft sign, etc.

Until Section~\ref{sec:full}, we will focus on this bottleneck problem (Eq.~\ref{eq:middle}). We will show a residual solver for this problem. Then we will unfold this iterative solver into a neural network, which is an enhanced version of the iterative solver. In Section~\ref{sec:full}, we will show how to embed this neural network into a larger network for the general TV regularized models.

\section{Residual Solver}
In this paper, we modify above iteration from~\cite{jia2010fast}, leading to a faster iterative algorithm. In FastSolver, during each iteration, $U^t$ has to be explicitly computed by Eq.~\ref{eq:jiaU} since it is needed in Eq.~\ref{eq:jiab}. Explicitly computing $U$ and its gradient adds extra computation burden for this algorithm. As shown in the following section, this computation can be avoided by finding the residual between the input and the optimal solution. Our method does not explicitly compute $U$ or its gradient. %It is an implicit solver.

Different from previous approaches, we propose a residual representation of Eq.~\ref{eq:middle} and perform the optimization in gradient domain. Moreover, our method does not explicitly compute $U$ during the optimization procedure. 

Specifically, from Eq.~\ref{eq:jiaU} we have
\begin{equation}
	U^{in}=V^{t+1}-\lambda\beta(\nabla^T\vec{b}^{in})\,.
\end{equation} Taking above equation into Eq.~\ref{eq:jiab}, we get
\begin{equation}
\label{eq:our}
\vec{b}^{in+1}=cut(\nabla V^{t+1}-\lambda\beta\nabla\nabla^T\vec{b}^{in}+\vec{b}^{in},\beta)\,.
\end{equation}
For a given $\nabla V^{t+1}$, the optimal solution $\vec{b}^*$ is a fixed point for this equation. After finding $\vec{b}^*$, the optimal solution $U^{t+1}$ can be obtained by Eq.~\ref{eq:jiaU}
\begin{equation}
\label{eq:residual}
U^{t+1}=V^{t+1}+(-\lambda\beta\nabla^T\vec{b}^*)\,.
\end{equation} In Eq.~\ref{eq:residual}, $-\lambda\beta\nabla^T\vec{b}^*$ is the residual between input $V^{t+1}$ and its optimal solution $U^*$. Since only the residual is solved, we call our method as Residual Solver (RS). And the residual term is defined as 
\begin{equation}
\label{eq:residualR}
R=-\lambda\beta\nabla^T\vec{b}^*\,.
\end{equation} Clearly, the residual $R$ is unique if $U^{t+1}$ is unique for given $V^{t+1}$. This algorithm is summarized in Algorithm~\ref{algo}. 

Our RS is computationally efficient. In RS, during the iteration in Eq.~\ref{eq:our}, only one vector field $\vec{b}$ has to be updated. And the two computation operations, $cut$ and $I-\lambda\beta\nabla\nabla^T$ can be efficiently performed, making RS faster. This is numerically confirmed in later section. 

\begin{algorithm}
	\caption{Residual Solver} \label{algo}
	\begin{algorithmic}[1]
		\Require input $V^{t+1}$, initial $\vec{b}^0=0$, $\beta$, Iteration $N$
		\State compute $\nabla V^{t+1}$
		\For {$in=0$ to $N-1$}, 
			 $$\vec{b}^{in+1}=cut(\nabla V^{t+1}+(1-\lambda\beta\nabla\nabla^T)\vec{b}^{in},\beta)$$
	    \EndFor
		\State	 $U^{t+1}=V^{t+1}+(-\lambda\beta\nabla^T\vec{b}^N)$
			 
	 \Ensure	output $U^{t+1}$
	\end{algorithmic}
\end{algorithm}

\subsection{Uniqueness of $\vec{b}$}
To show that our RS is numerical stable, we need to find a unique vector field $\vec{b}$ for the unique residual $R$. It is well-known that Eq.~\ref{eq:middle} is convex~\cite{TV1992}. Therefore, its optimal solution $U^{t+1}$ is unique, leading to the uniqueness of the residual $R=-\lambda\beta\nabla^T\vec{b}^*$. Although the vector field $\vec{b}^*$ is not unique, it belongs to a unique set in terms of divergence $\nabla^T$ equivalency
\begin{equation}
{\cal S}=\left\{\vec{b}~|~\nabla^T\vec{b}=\nabla^T\vec{b}^*\,,\|\vec{b}\|_2<+\infty\right\}\,.
\end{equation}

This set $\cal S$ is convex because $\nabla^T$ is a linear operation. Let $0\le\beta\le1$ be a scalar. Then, for any $\vec{b}_1$ and $\vec{b}_2$ in this set, since $\nabla^T(\beta\vec{b}_1+(1-\beta)\vec{b}_2)=\nabla^T\vec{b}^*$, we have
\begin{equation}
\nabla^T(\beta\vec{b}_1+(1-\beta)\vec{b}_2)\in{\cal S}=	\left\{\vec{b}~|~\nabla^T\vec{b}=\nabla^T\vec{b}^*\right\}\,.
\end{equation}This result indicates that this unique set is a convex set. 

Since the ground truth image $U$ has bounded intensity values (usually $U(\vec{x})\in[0,1])$ or $U(\vec{x})\in[0,255])$), $\nabla^T\vec{b}^*$ is also bounded ($<+\infty$). Therefore, we can require $\|\vec{b}\|_2<+\infty\,,\forall \vec{b}\in {\cal S}$.

Thank to the bounded norm and convexity, we can obtain a unique solution if we add an additional constraint, such as minimizing $||\vec{b}||_1$. More specifically,  we have
\begin{equation}
\vec{b}_1^*=\arg\min_{\vec{b}}||\vec{b}||_1,~s.t.~\vec{b}\in{\cal S}\,.
\end{equation} Thanks to the convexity of $\cal S$, $\vec{b}_1^*$ is unique.

In practice, we always initialize $\vec{b}^0=0$. This initialization leads to the unique $\vec{b}_1^*$ since it is the ``minimum'' vector field in this convex set. We observed that the same input always find the same vector field $\vec{b}_1^*$ in practice. 

The convexity of the solution set and uniqueness of vector field $\vec{b}_1^*$ provide the theoretical guarantees for our RS. They also reveal the numerical stability of our RS, which is confirmed in following experiment on BSDS500 data set.

\subsection{Convergence to the Optimal Solution}
\begin{figure}[!tbh]
	\subfigure[$E_{RS} - E_{FS}$ distribution]{		
		\includegraphics[width=0.48\linewidth]{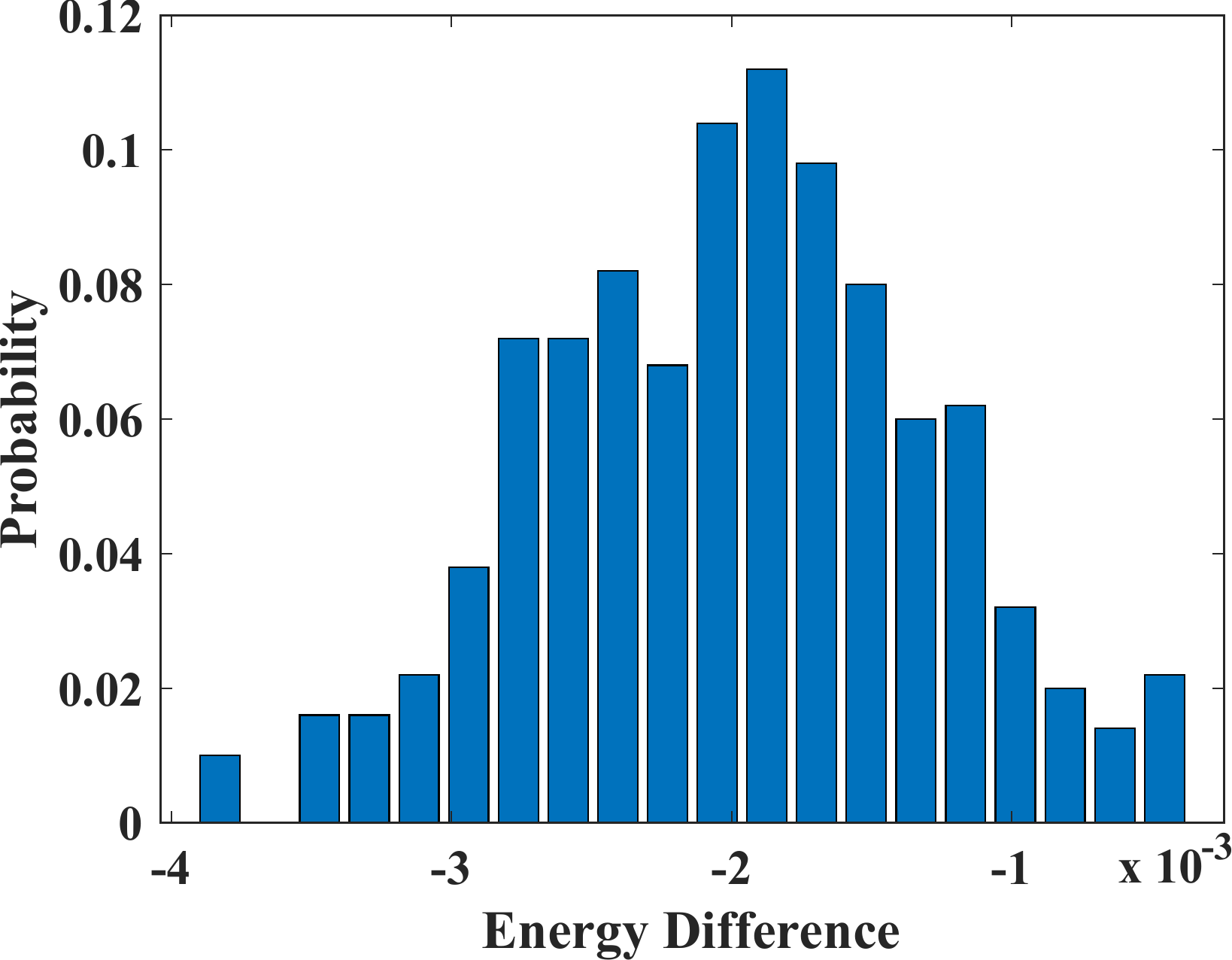}	
		\subfigure[$SSIM_{RS}-SSIM_{FS}$ ]{		
			\includegraphics[width=0.48\linewidth]{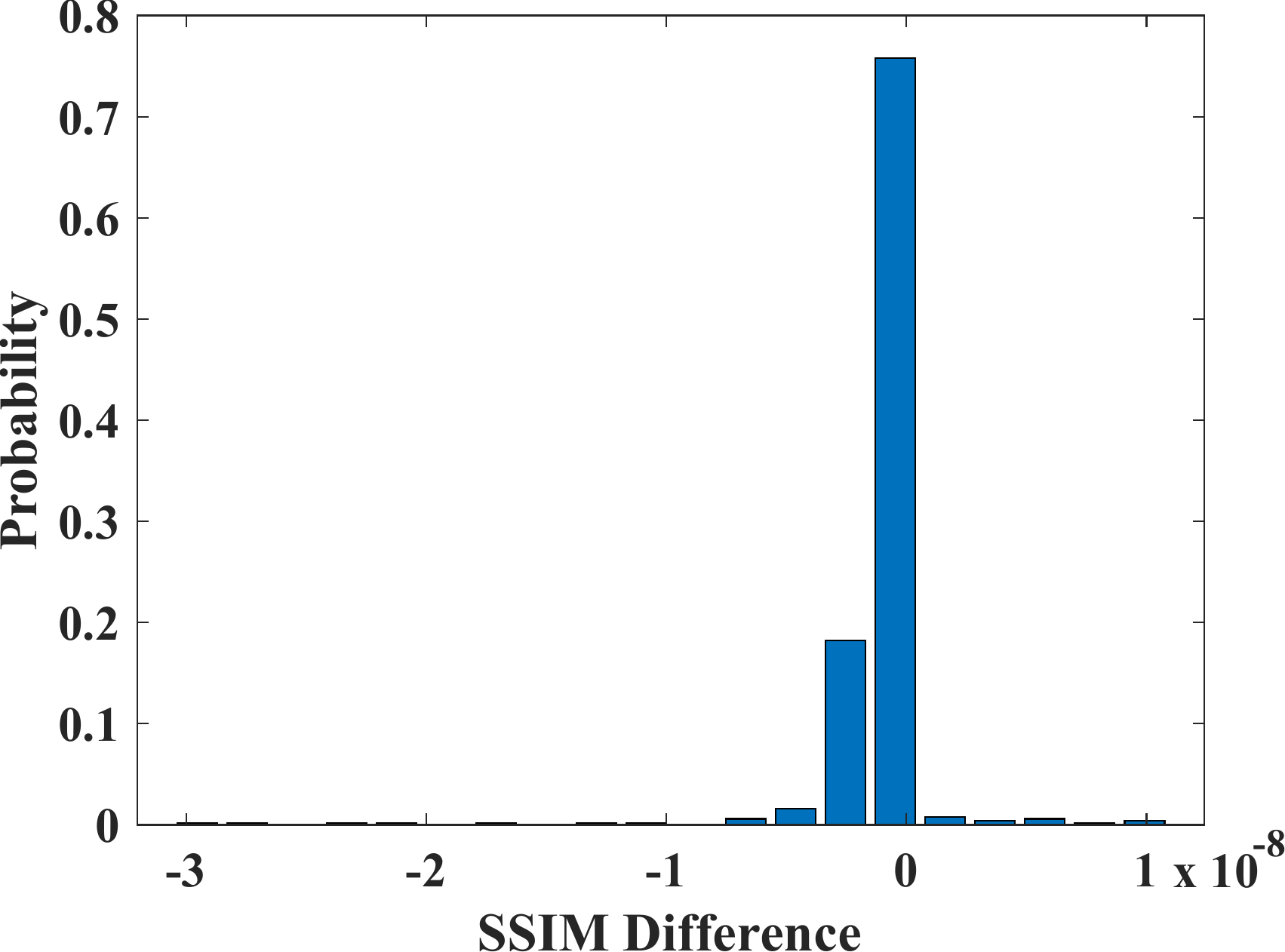}}}
	\caption{Comparison of FastSolver and our Residual Solver on 500 natural images. The distribution of $E_{RS} - E_{FS}$ is shown in (a), where the negative value indicates that RS can reach lower energy. The distribution of ssim difference is shown in (b). Be aware that the difference is about $10^{-8}$. \label{fig:energy1} }
\end{figure}

Our RS can reach the optimal solution. We run RS and FS with 200 iterations on 500 natural images from BSDS500 data set. Then we compute their final loss function value for each image, respectively. For the same image, we compute the loss function value difference. The distribution of such differences for all 500 images is shown in Fig.~\ref{fig:energy1} (a). The average difference is about $-2\times10^{-3}$. The negative values indicate that our residual solver always reaches lower energy level for these images. The results confirm that RS can reach the optimal solution. 

We also compute the SSIM between the original and processed images. For FastSolver and RS, the SSIM difference is small ($\approx10^{-8}$) and its distribution is shown in Fig.~\ref{fig:energy1} (b). 

In addition, our RS runs faster. The average running time of RS and FastSolver for the 500 color images on the same hardware with the same iteration number is 0.89 seconds and 1.07 seconds, respectively. This indicates that our RS can reduce about $17\%$ of the running time.

\subsection{Properties}
This new residual solver has three important properties: 
\begin{itemize}
	\item {\bf implicit}: Our RS does not require to compute $U^t$ during the iteration. In contrast, $U^t$ has to be explicitly computed in Eq.~\ref{eq:jiaU} for each iteration in FastSolver. 
	\item {\bf gradient domain}: Our RS solves the problem in the gradient domain, which is different from previous methods that work in the intensity domain.
	\item {\bf fast}: In Eq.~\ref{eq:our}, $\nabla V^{t+1}$ is fixed during the inner iterations. Therefore, $\nabla V^{t+1}$ can be computed before the loop, which reduces about $17\%$ running time in practice. 
\end{itemize}

\section{Residual Solver Network}
In this paper, we propose a new network, based on the residual solver in the previous section. Inspired by the neural network development in the past few years (Section~\ref{sec:nn}), we unfold the RS algorithm into a neural network, called RSnet, which can be considered as an ``enhanced'' version of our iterative algorithm. 

The enhancement comes from two aspects. First, there are only two components $\vec{b}=(b_x, b_y)$ in RS, covering the $x$ and $y$ axis direction respectively. In contrast, RSnet uses more channels (eight, sixteen or thirty two), which cover more orientations (including the $x$ and $y$ directions). Second, the operations (gradient and divergence) in RS are independent from the input. In contrast, these operations in RSnet are learned from training data set and thus more adaptive in solving this problem. Therefore, RS iterative algorithm can be considered as a special case of RSnet with two channels and fixed convolution kernels. And RSnet is an ``enhanced'' version of RS.

\begin{figure}[!t]
	\begin{center}
		\includegraphics[width=\linewidth]{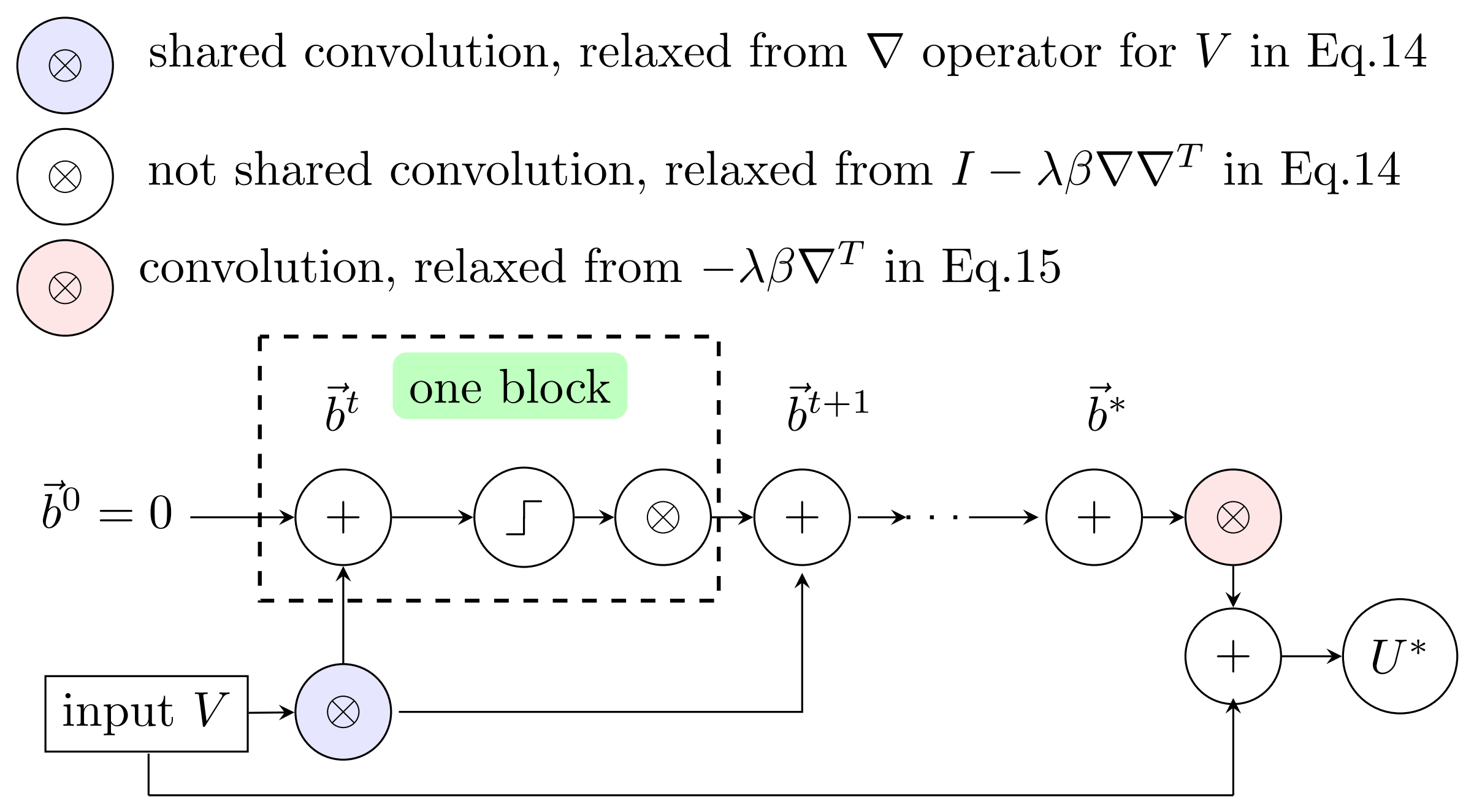}
	\end{center}
	\caption{The network architecture of RSnet, where $\otimes$ indicates learnable convolution kernels and the step function indicates the $cut$ or $clip$ activation function. RSnet does {\it not} contain any batch normalization or drop out layers. Therefore, it can be trained on low resolution images but applied on high resolution images.}
	\label{fig:net}
\end{figure} 
\subsection{Network Architecture}
First, the gradient operator for $V$ is relaxed to a convolution kernel in our neural network. And this convolution kernel is shared among each unfolding block. Second, the linear operator $I-\lambda\beta\nabla\nabla^T$ ($I$ is the identity matrix) for $\vec{b}^t$ in Eq.~\ref{eq:our} is relaxed to a convolution kernel. But it is not shared for the network blocks in order to improve the neural network's capacity. Finally, the $-\lambda\beta\nabla^T$ operator in Eq.~\ref{eq:residual} is relaxed to a convolution kernel. The network architecture is shown in Fig.~\ref{fig:net}. One block corresponds to one iteration in RS.

In this network, $V$ is the input for our neural network. The convolution of $V$ is shared for all blocks, which is shown in shaded blue. The step symbol indicates the $cut$ activation function, which is implemented by clip operation. The $\otimes$ indicates a convolution kernel that is not shared between blocks. The initial $\vec{b}^0=0$ is the same as the iterative algorithm. Different from previous related neural networks, our network does not adopt any Batch Normalization. Therefore, it can be trained on low resolution images but applied on high resolution images.

The proposed RSnet has three important features:
\begin{itemize}
	\item Our network is unsupervised. It does not require the ground truth images. It is suitable for practical applications where the ground truth is unknown, such as biomedical image reconstruction.
	\item Our network can be trained on low resolution images but applied on high resolution images. This property is important for training process on hardware with limited memory. Our designed loss function is independent from image resolution and our network does not contain any batch normalization nor drop out layer. %This property will be confirmed in the experiment section.
	\item The result from our network is comparable with the counterpart from classical iterative algorithms. As shown in Section~\ref{sec:ourloss}, we can set the parameter in our loss function for this network such that results from our method are comparable with the counterpart from classical iterative algorithms.
\end{itemize}

\subsection{The Loss Function}
\label{sec:ourloss}
Here, we propose a loss function that is independent from the image resolution. Although Eq.~\ref{eq:loss} is well-known from the literature, it depends on the resolution of input image because the $\lambda$ is for the whole image rather than for each pixel. To relax this constraint, we propose to use following loss function (normalized version) for our neural network
\begin{equation}
\label{eq:loss2}
\min_{U}\left\{\frac{\frac{1}{2}\|AU(\vec{x})-f(\vec{x})\|_2^2}{\int_{\vec{x}}1\mathrm{d}\vec{x}}+{\lambda}\frac{TV(U(\vec{x}))}{\int_{\vec{x}}1\mathrm{d}\vec{x}} \right\}\,.
\end{equation} 

This loss function is independent from the image resolution. Therefore, this loss function is comparable for different resolution images. In contrast, Eq.~\ref{eq:loss} is NOT comparable for images with different resolution. When we compare the energy levels between iterative algorithms and our neural network method, we simply scale them by the total number of pixels ($\int_{\vec{x}}1\mathrm{d}\vec{x}$). With this loss function and our network architecture, {\bf our neural network can be trained on low resolution images but applied on high resolution images}. This property is different from traditional neural networks that involve batch normalization layers or drop out layers. They can only deal with the same resolution images (training and inference images must have the same resolution).

\begin{figure}[!t]
	\begin{center}
		\includegraphics[width=0.8\linewidth]{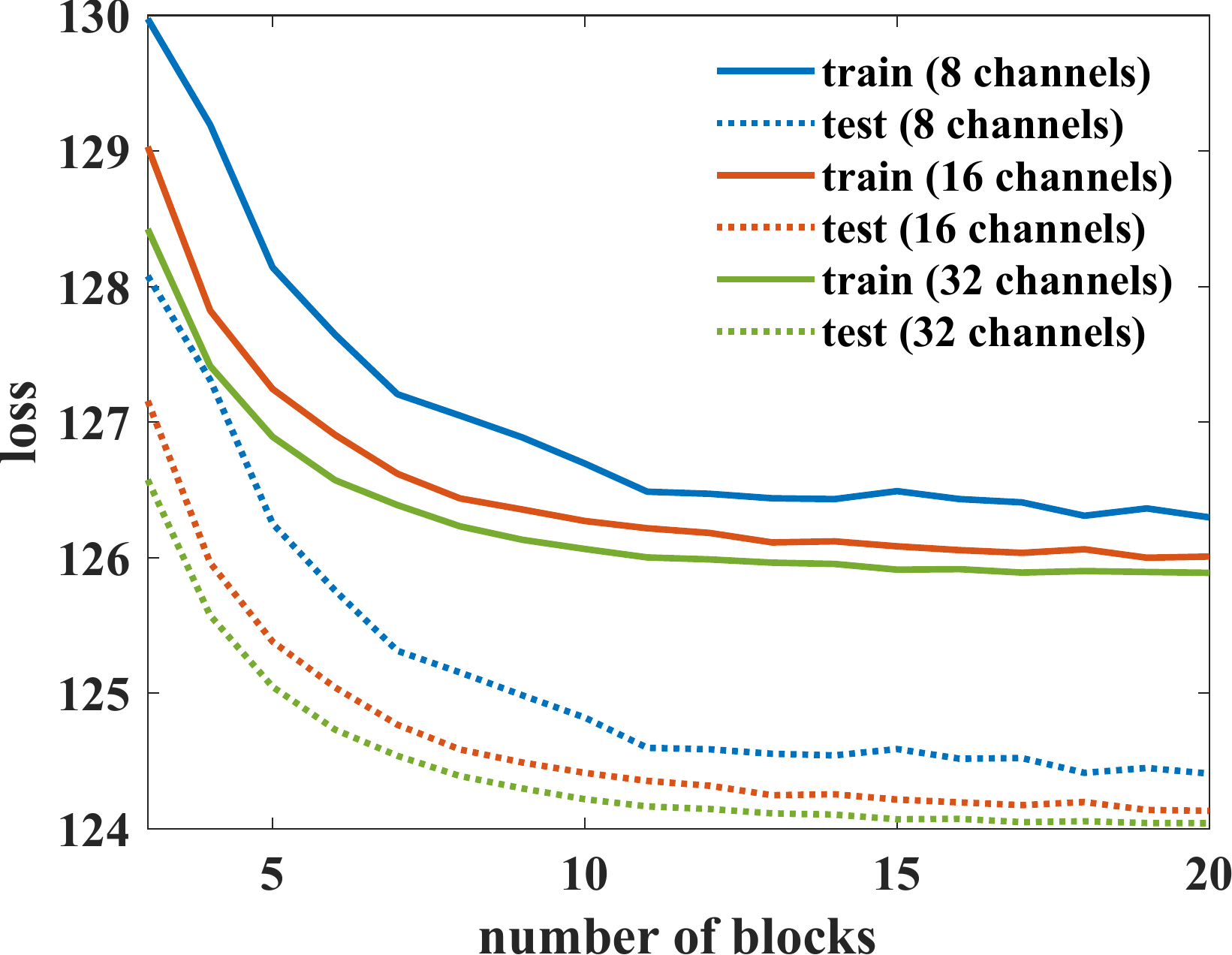}
	\end{center}
	\caption{Our network with different number of blocks and channels was trained with 3000 epochs on natural image patches. And the loss value is shown. As comparison, the loss function value from the classical FastSolver on the same data set is 126.}
	\label{fig:lossconverge}
\end{figure}

\begin{figure}[!t]
	\begin{center}
		\subfigure[$E_{RSnet} - E_{FS}$ distribution]{		
			\includegraphics[width=0.48\linewidth]{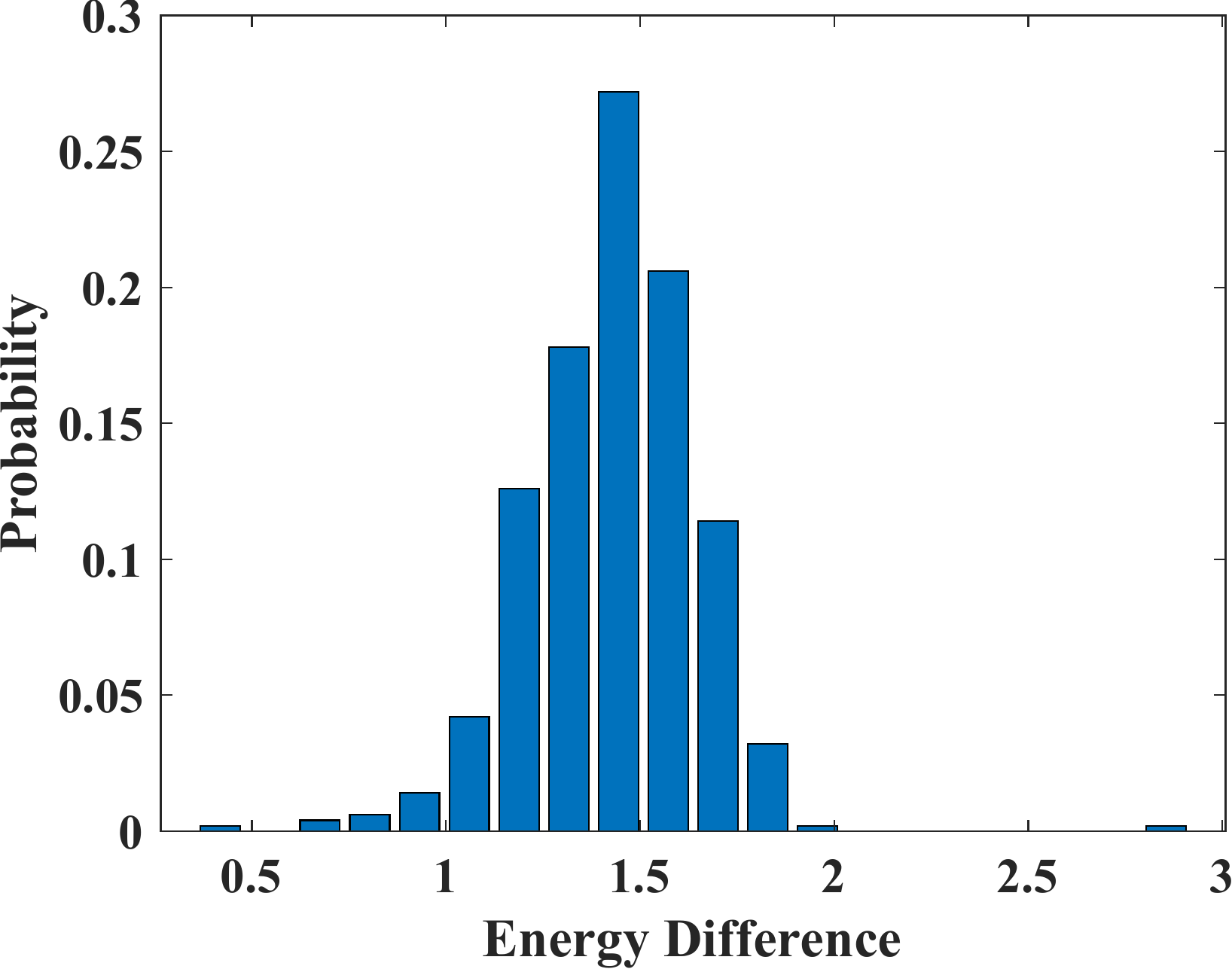}}		
		\subfigure[$SSIM_{RSnet}-SSIM_{FS}$ ]{		
			\includegraphics[width=0.48\linewidth]{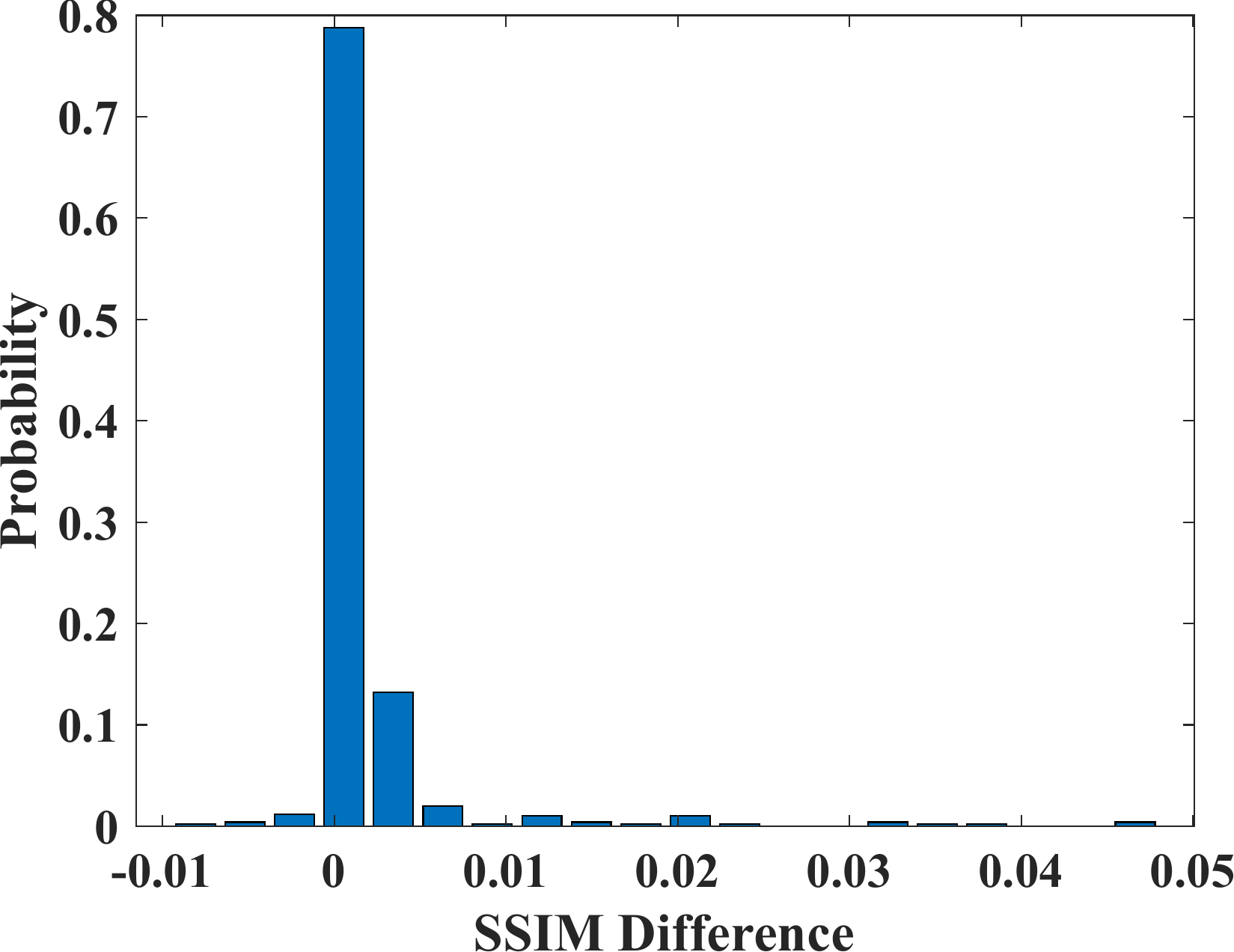}}
	\end{center}
	\caption{Comparison of FastSolver and RSnet on 500 natural images. For each image, the energy and SSIM is computed. The distribution of their difference is shown.}
	\label{fig:imagesmoothingperf}
\end{figure}

\begin{figure*}[!tbh]
	\begin{center}
		\subfigure[kernels in the first layer]{
			\includegraphics[width=0.315\linewidth]{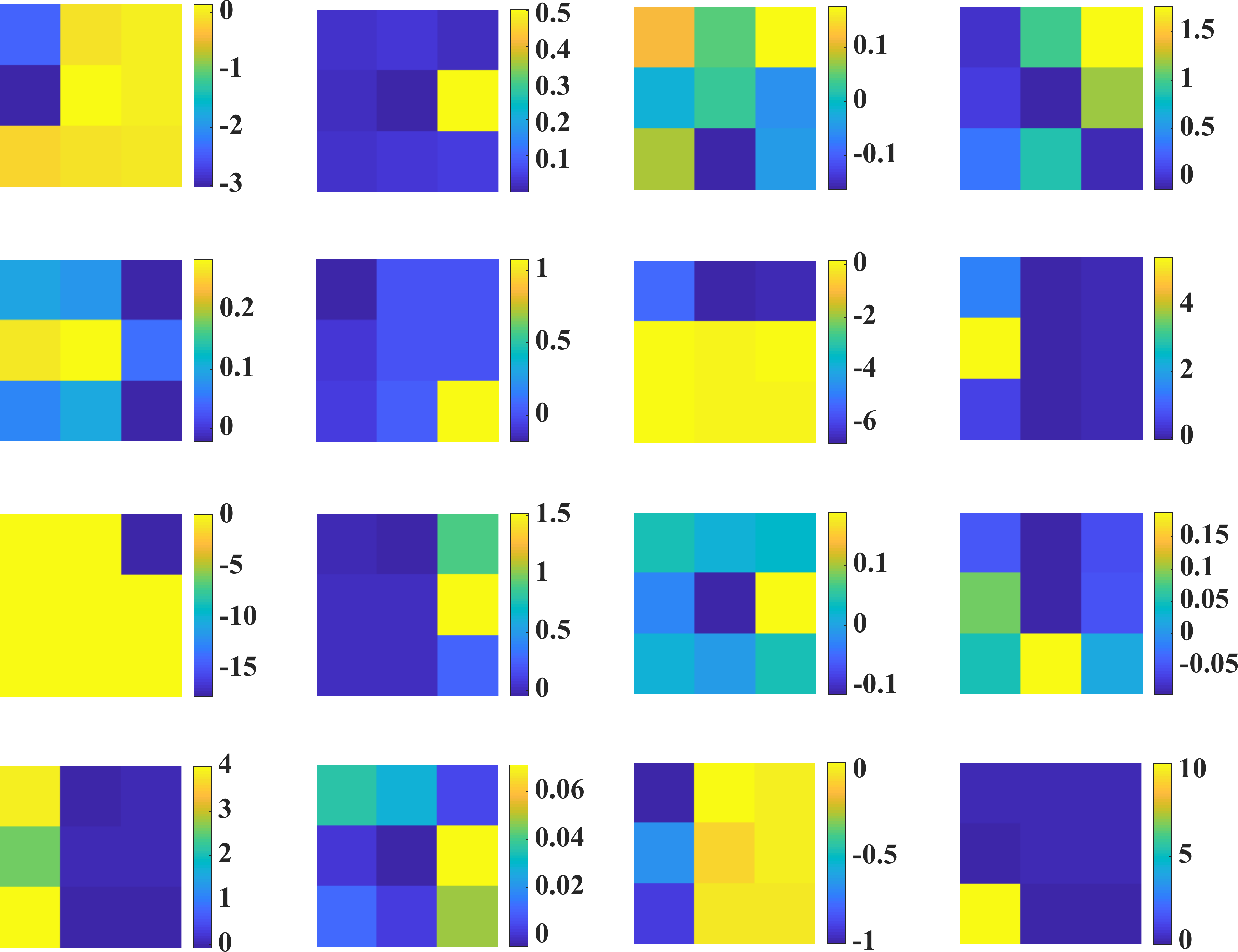}}
		\subfigure[kernels (shared layer)]{
			\includegraphics[width=0.315\linewidth]{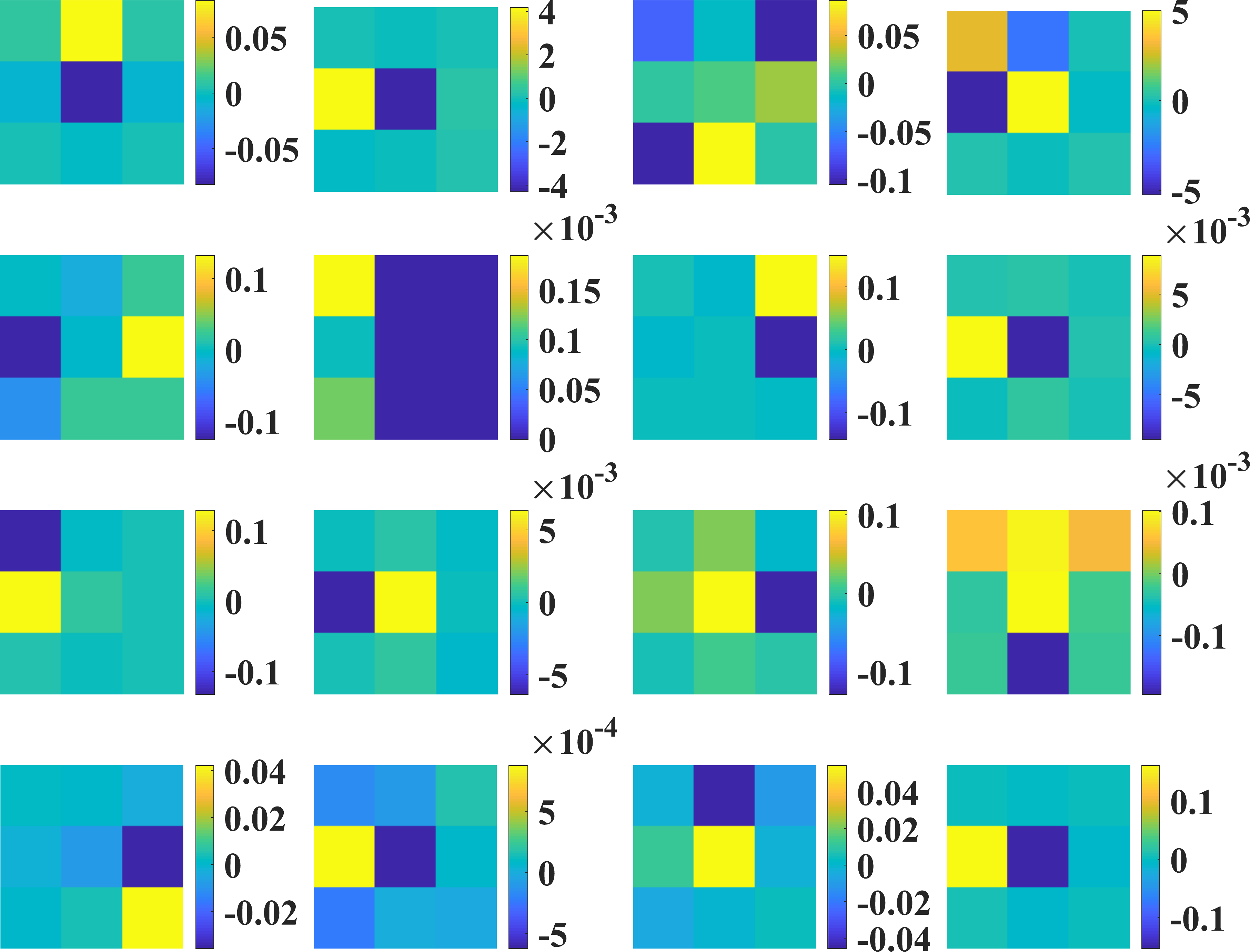}}
		\subfigure[kernels (last layer)]{
			\includegraphics[width=0.315\linewidth]{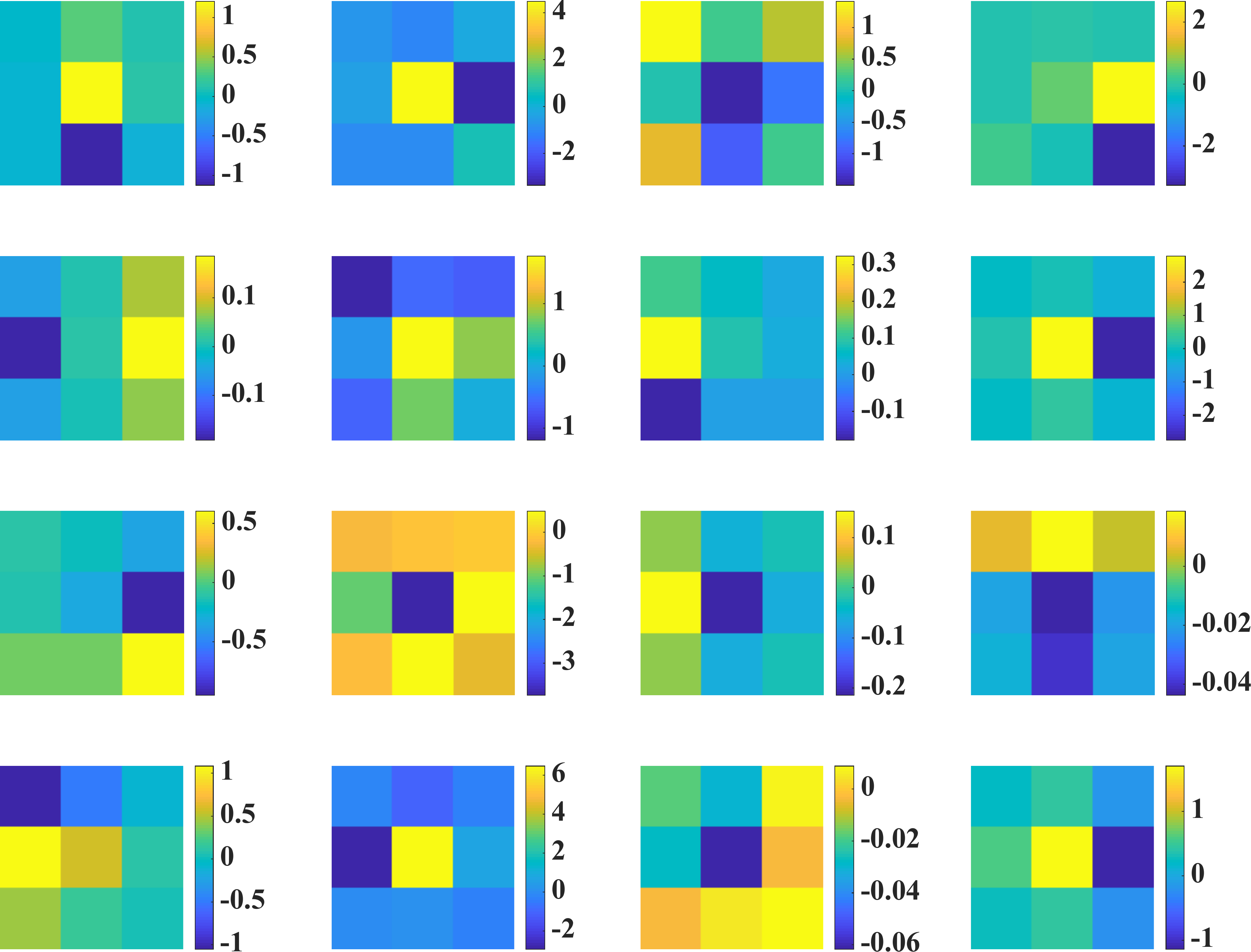}}	
	\end{center}
	\caption{The 16 learned $3\times 3$ kernels in RSnet. These kernels cover the classical gradient and divergence operators. RSnet is an enhanced version of RS.}
	\label{fig:kernels}
\end{figure*}

\subsection{Training on Natural Images}
We randomly crop 10,000 image patches with $128\times 128$ resolution from 500 natural images in the BSDS dataset~\cite{bsds}. Each patch is turned into gray scale. Eighty percent of these patches are used as training set and the rest is the validation set. During training, we fix ${\lambda}=10$, all kernel size is set to $3\times 3$, learning rate is 0.0003, batch size is 64, and the epoch number is 3000. We increase the number of block from 3 to 20 and set the number of channels as 8, 16 and 32, respectively.

After training, the final loss function values are shown in Fig.~\ref{fig:lossconverge}. When increasing the number of blocks or the number of channels, the loss is reduced after training. After some value, increasing the them does not improve the result anymore. As comparison, the loss function value for the optimal solutions from FastSolver is 126. This result shows that our neural network can reach the optimal results.

Keeping in mind the complexity of network and the effectiveness, we fix the number of block as 10 and the number of channel as 16 in the rest of our paper. This configuration leads to 21K learnable parameters. Such small number of parameters can be easily trained with a modern GPU hardware.  

The loss function values are compared with the counterpart from FastSolver. The distribution of their difference is shown in Fig.~\ref{fig:imagesmoothingperf} (a). We further compare their SSIM difference and the distribution is shown in Fig.~\ref{fig:imagesmoothingperf} (b). RSnet keeps more structural information (Fig.~\ref{fig:imagesmoothingperf} (b)).

Some learned convolution kernels are visualized in Fig.~\ref{fig:kernels}. These kernels already contain the classical gradient and divergence operators. Therefore, this network is an enhanced version of our iterative algorithm that only has gradient and divergence operators. As a result, this network has better performance in some applications (Section~\ref{sec:denoise}).

RSnet takes only 0.064 second for each color image with 481$\times$321 resolution from BSDS500 data set on a GTX 1080 Ti GPU card. In contrast, the average running time of RS and FastSolver is 0.89 seconds and 1.07 seconds, respectively. 

\subsection{Embedding RSnet into Other Networks}
\label{sec:full}
We have shown the residual solver and its unfolding RSnet that can efficiently solve the bottleneck problem in TV regularized models.

\begin{figure}[!t]
	\begin{center}
		\includegraphics[width=0.65\linewidth]{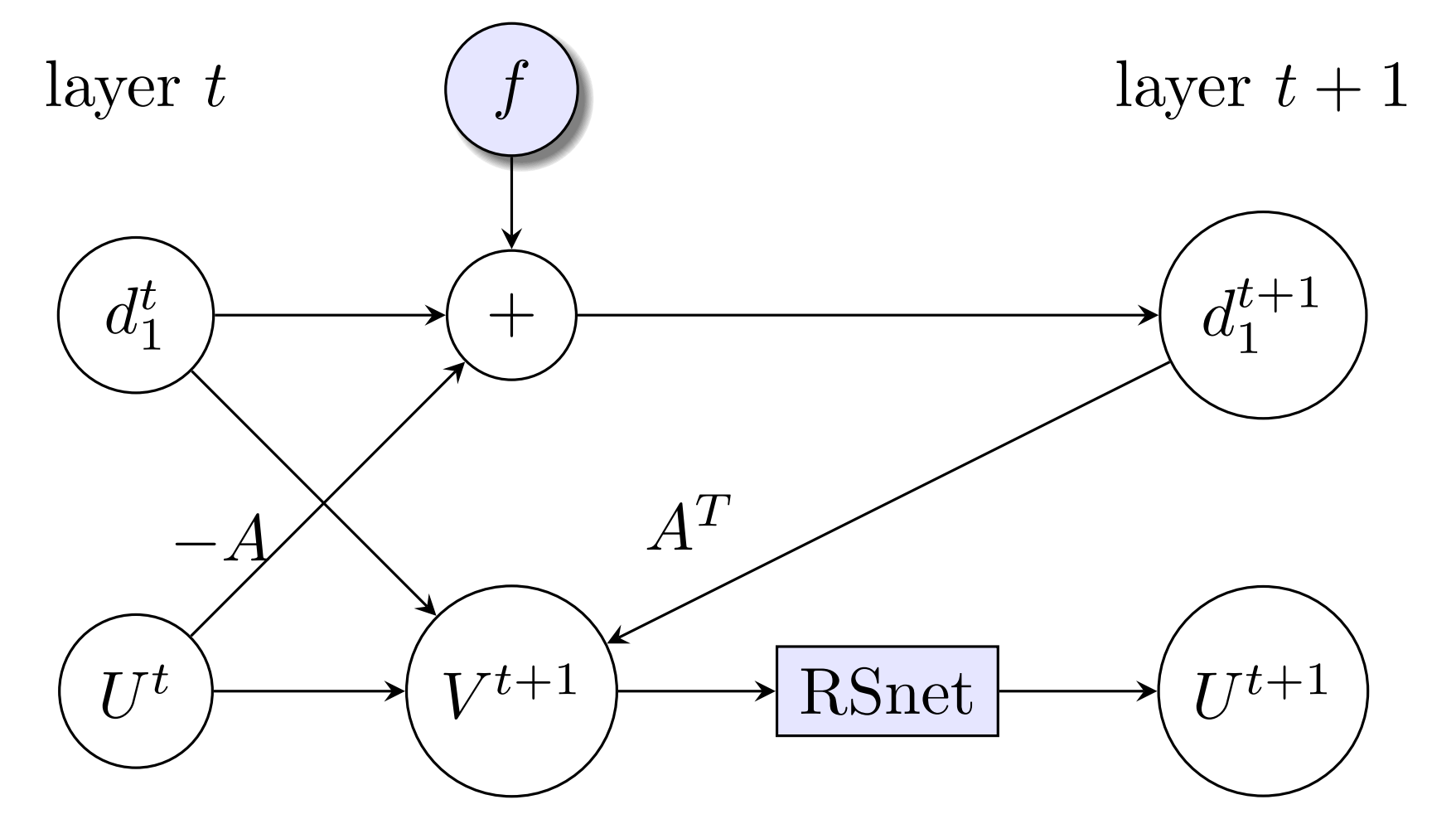}
	\end{center}
	\caption{Our RSnet can be embedded in the full network to solve the general TV regularized models.}
	\label{fig:fullNet}
\end{figure}

In this section, we show how to embed RSnet into a larger network that can solve the general total variation regularized models (Eq.~\ref{eq:loss}). More specifically, we unfold the dual first iteration algorithm (Eq.~\ref{eq:dual} to Eq.~\ref{eq:middle}) into a neural network, which contains RSnet as a sub network for Eq.~\ref{eq:middle}. The network structure is shown in Fig.~\ref{fig:fullNet}.

\section{Applications}
In this section, the proposed RS and RSnet are applied in three scenarios, image smoothing, image denoising and ultrasonic image reconstruction. In image smoothing, we show that RS and RSnet can reach the optimal solution as the classical FastSolver~\cite{jia2010fast}. This experiment also confirms that RSnet can be trained on low resolution images ($128\times 128$) but used on other resolution images ($321\times 481$ in this case). In image denoising, we show that RSnet can get better results because it is an enhanced version of RS. In this scenario, we directly adopt the already trained RSnet from the image smoothing problem and apply it on noisy images, showing the transferring ability of RSnet. In the ultrasonic image reconstruction, we show that RS and RSnet can be used for a global operation, not limited in local operations, such as smoothing and denoising. In all these scenarios, we compared the RS and RSnet results with the counterpart from the FastSolver~\cite{jia2010fast}. In all these applications, we use Eq.~\ref{eq:loss2} as the loss function and set ${\lambda}=10$. All the experiments are performed on a GTX 1080 Ti GPU.
\subsection{Image smoothing}
\label{sec:smooth}

In image smoothing, we show that RS and RSnet can reach the optimal solution as FastSolver~\cite{jia2010fast}. This experiment also confirms that RSnet can be trained on low resolution images but applied on other resolution images. 

\begin{figure*}[!bht]
	\begin{center}
		\subfigure[Input]{
			\includegraphics[width=0.23\linewidth]{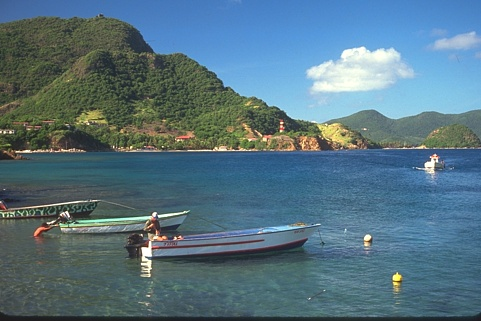}}
		\subfigure[FS (SSIM:0.88)]{
			\includegraphics[width=0.23\linewidth]{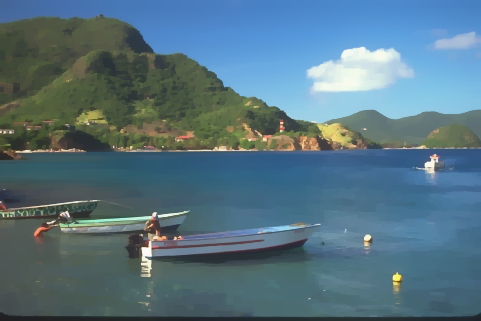}}
		\subfigure[RS (SSIM:0.88)]{
			\includegraphics[width=0.23\linewidth]{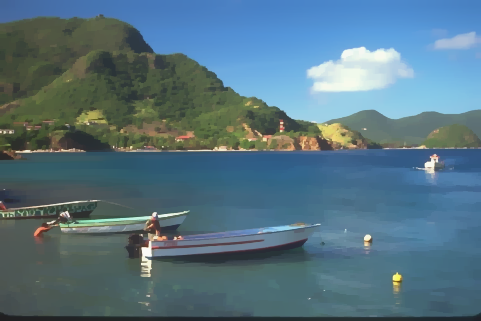}}
		\subfigure[{\scriptsize RSnet(ssim:0.89)}]{
			\includegraphics[width=0.23\linewidth]{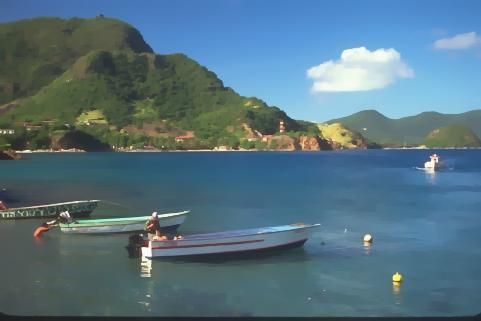}}\\
		\subfigure[Input patch]{
			\includegraphics[width=0.23\linewidth]{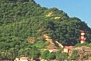}}
		\subfigure[FS patch]{
			\includegraphics[width=0.23\linewidth]{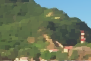}}
		\subfigure[RS patch]{
			\includegraphics[width=0.23\linewidth]{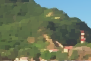}}
		\subfigure[RSnet patch]{
			\includegraphics[width=0.23\linewidth]{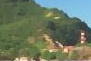}}\\
		\subfigure[Input]{
			\includegraphics[width=0.23\linewidth]{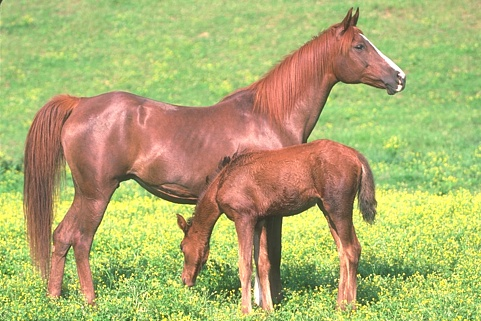}}
		\subfigure[FS (SSIM:0.96)]{
			\includegraphics[width=0.23\linewidth]{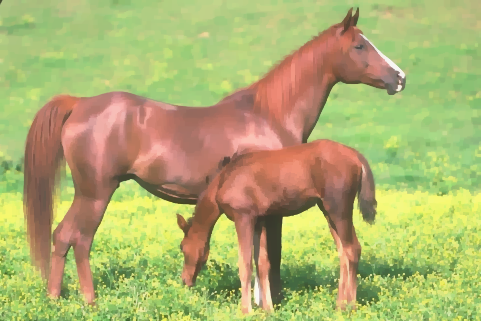}}
		\subfigure[RS (SSIM:0.96)]{
			\includegraphics[width=0.23\linewidth]{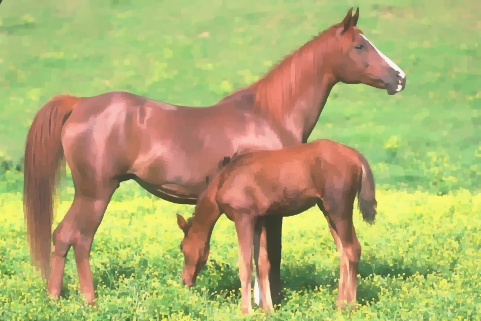}}
		\subfigure[{\scriptsize RSnet (SSIM:0.96)}]{
			\includegraphics[width=0.23\linewidth]{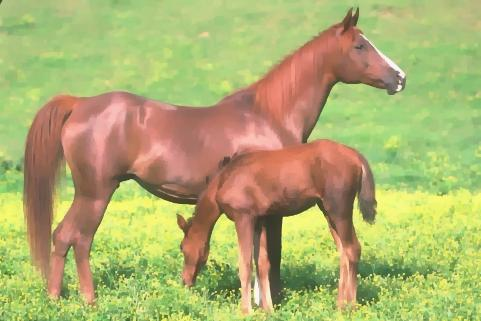}}\\
		\subfigure[Input patch]{
			\includegraphics[width=0.23\linewidth]{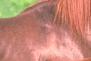}}
		\subfigure[FS patch]{
			\includegraphics[width=0.23\linewidth]{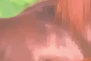}}
		\subfigure[RS patch]{
			\includegraphics[width=0.23\linewidth]{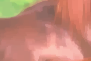}}
		\subfigure[RSnet patch]{
			\includegraphics[width=0.23\linewidth]{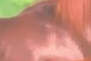}}		
	\end{center}
	\caption{Two examples from BSDS500 for the image smoothing problem. These three methods achieve similar results.}
	\label{fig:imagesmoothing}
\end{figure*}

In image smoothing, the imaging matrix $A$ is simply the identity matrix. We apply FastSolver, RS or RSnet on 500 color images from BSDS dataset. And the algorithms perform on each color channel separately. The iteration number of FastSolver and RS is set to 200. The RSnet is trained on 10,000 image patches with resolution $128\times128$, but applied on $321\times481$ or $481\times321$ resolution images. Two examples from the data set are shown in Fig.~\ref{fig:imagesmoothing}. As can be told from Fig.~\ref{fig:imagesmoothing}, the smoothing results are comparable. To evaluate the quality of these results, structural similarity index measure (SSIM) is used to calculate the similarity between the input and the output images~\cite{ssim}. 

\begin{figure}[!tbh]
	\begin{center}	
		\subfigure[Fig.~\ref{fig:imagesmoothing}(b) vs. (c)]{		
			\includegraphics[width=0.48\linewidth]{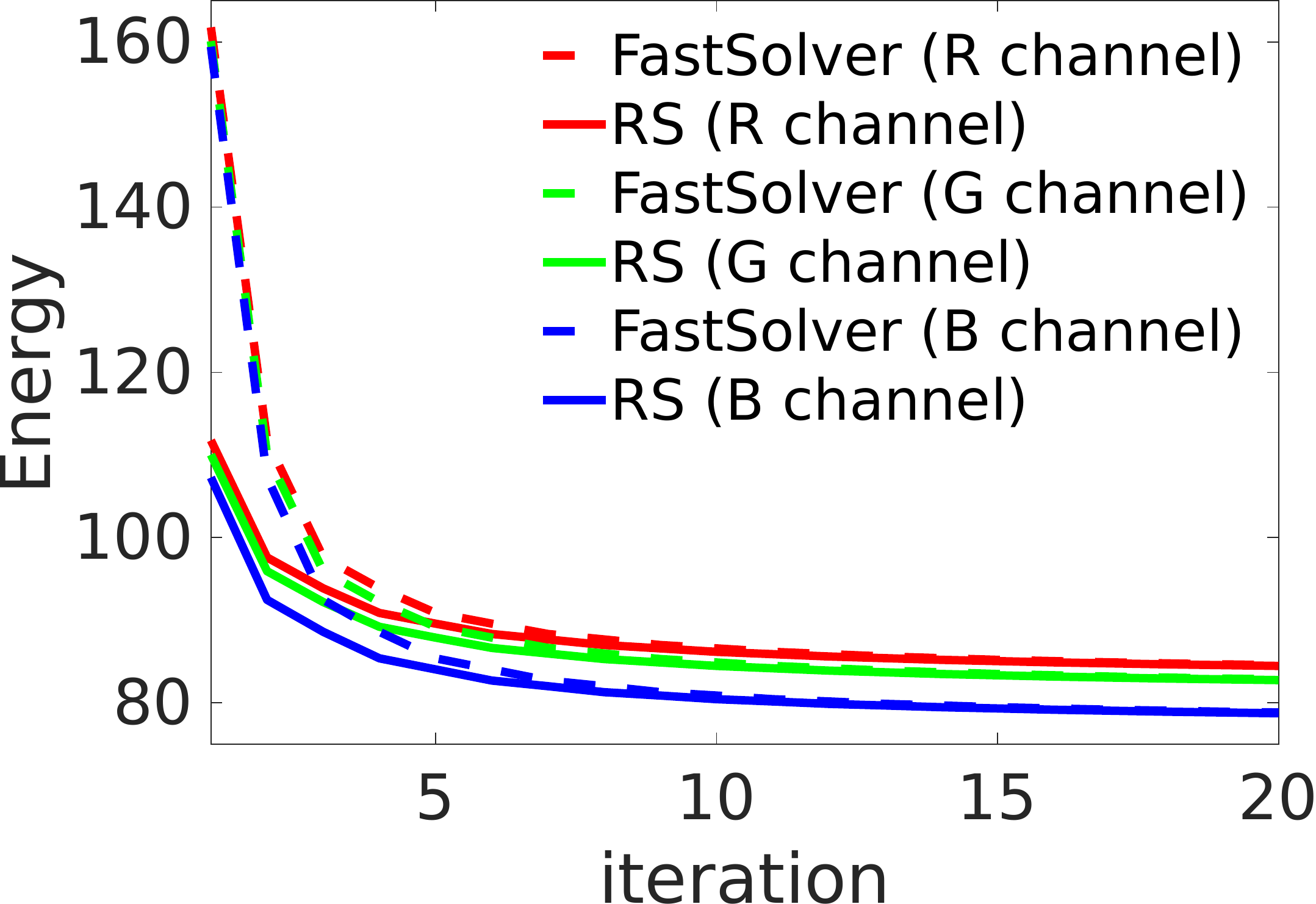}}	
		\subfigure[Fig.~\ref{fig:imagesmoothing}(j) vs. (k)]{
			\includegraphics[width=0.48\linewidth]{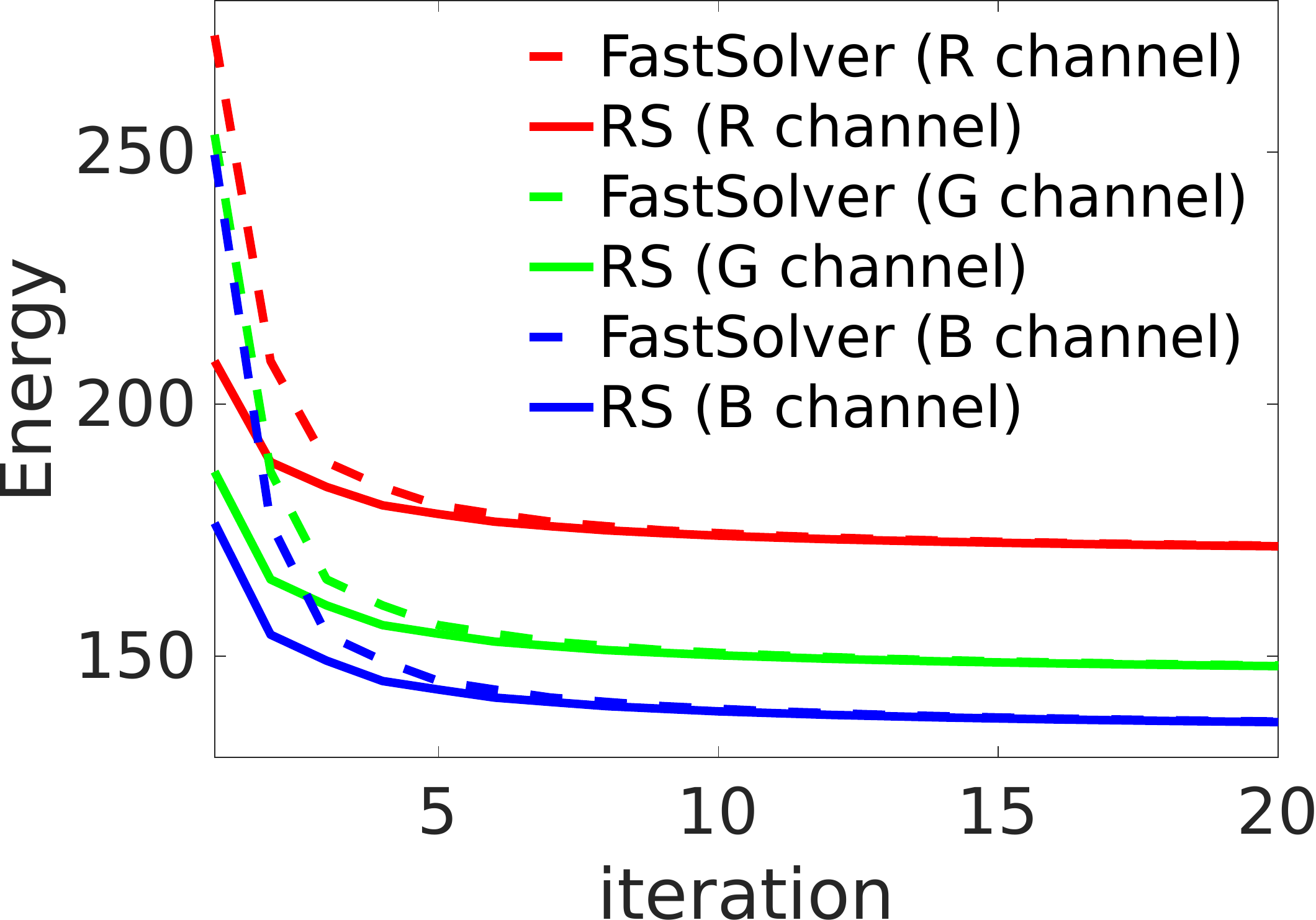}}	
	\end{center}
	\caption{The convergence of energy for FastSolver and RS.}
	\label{fig:imagesmoothingconverge}
\end{figure}

For these two examples, we also calculate the energy for both methods with first 20 iterations. The results are shown in Fig.~\ref{fig:imagesmoothingconverge}. RS converges faster than FastSolver at the first few iterations. We observed that this fact is also true for other images from the BSDS500 data set. This is because the operations in RS are independent from the image content.

\begin{figure*}[!tbh]
	\begin{center}
		\subfigure[Ground truth]{
			\includegraphics[width=0.18\linewidth]{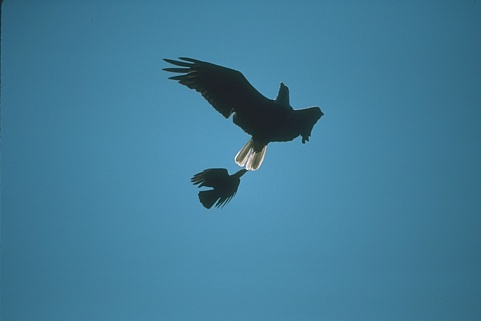}}
		\subfigure[noisy (PSNR:24.62)]{
			\includegraphics[width=0.18\linewidth]{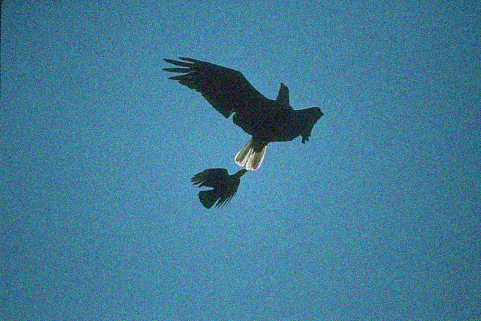}}
		\subfigure[FastSolver (PSNR:38.87)]{
			\includegraphics[width=0.18\linewidth]{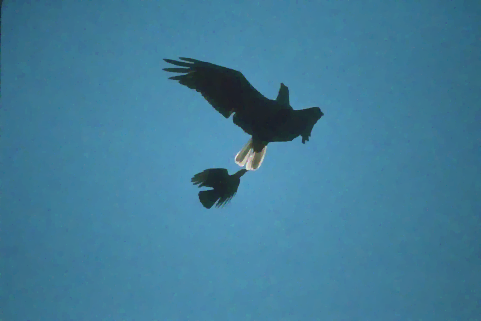}}
		\subfigure[RS (PSNR:38.87)]{		
			\includegraphics[width=0.18\linewidth]{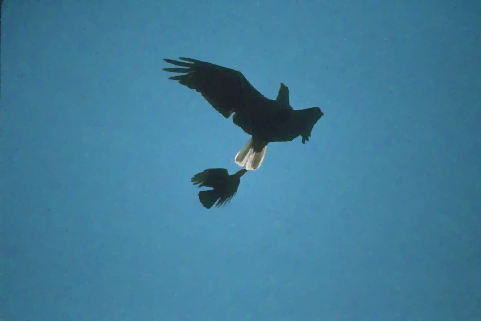}}
		\subfigure[RSnet (PSNR:42.32)]{
			\includegraphics[width=0.18\linewidth]{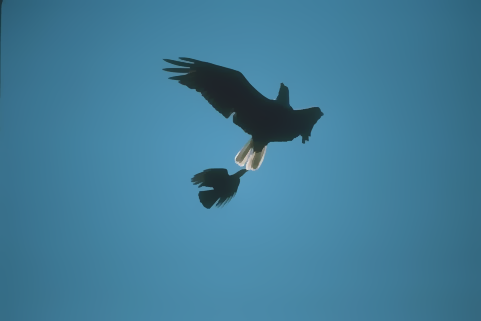}}\\
		\subfigure[Truth patch]{
			\includegraphics[width=0.18\linewidth]{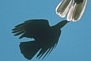}}
		\subfigure[noisy patch]{
			\includegraphics[width=0.18\linewidth]{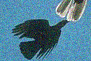}}
		\subfigure[FastSolver]{
			\includegraphics[width=0.18\linewidth]{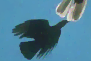}}
		\subfigure[RS patch]{		
			\includegraphics[width=0.18\linewidth]{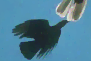}}
		\subfigure[RSnet patch]{
			\includegraphics[width=0.18\linewidth]{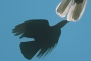}}\\
		\subfigure[Ground truth]{
			\includegraphics[width=0.18\linewidth]{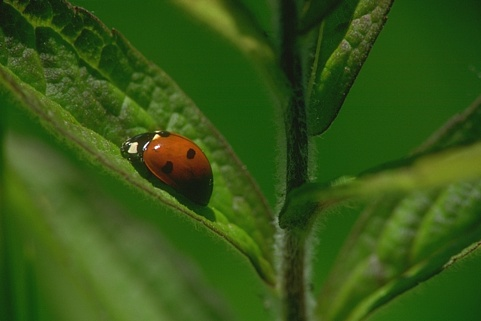}}
		\subfigure[noisy (PSNR:24.98)]{
			\includegraphics[width=0.18\linewidth]{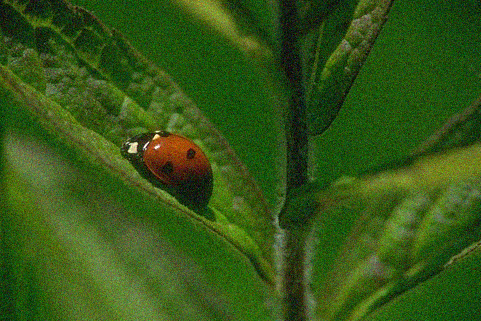}}
		\subfigure[FastSolver(PSNR:34.7)]{
			\includegraphics[width=0.18\linewidth]{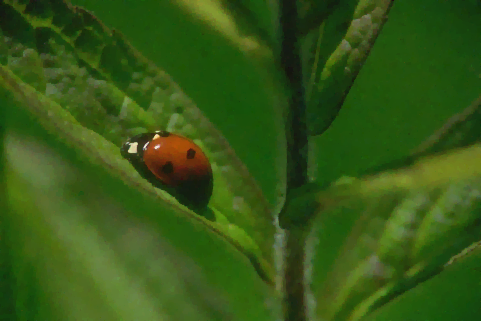}}
		\subfigure[RS (PSNR:34.76)]{
			\includegraphics[width=0.18\linewidth]{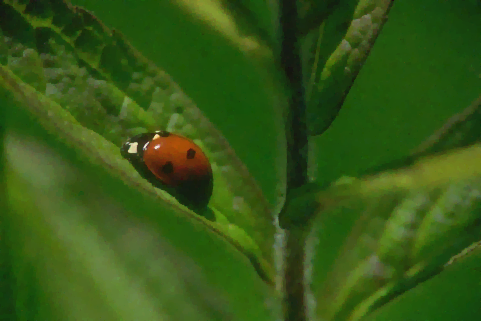}}
		\subfigure[RSnet (PSNR:36.31)]{
			\includegraphics[width=0.18\linewidth]{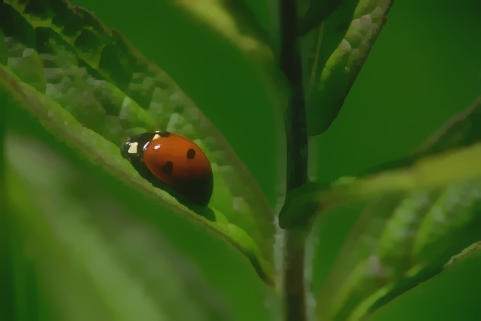}}\\
		\subfigure[Truth patch]{
			\includegraphics[width=0.18\linewidth]{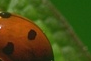}}
		\subfigure[noisy patch]{
			\includegraphics[width=0.18\linewidth]{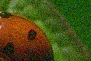}}
		\subfigure[FastSolver]{
			\includegraphics[width=0.18\linewidth]{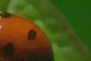}}
		\subfigure[RS patch]{
			\includegraphics[width=0.18\linewidth]{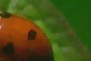}}
		\subfigure[RSnet patch]{
			\includegraphics[width=0.18\linewidth]{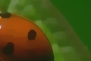}}	
	\end{center}
	\caption{Two examples and their zoomed patches from BSDS500 for the image denoising problem. }
	\label{fig:imagedenoising}
\end{figure*}

\begin{figure}[!tbh]
	\begin{center}
		\subfigure[$E_{RS} - E_{FS}$ distribution]{		
			\includegraphics[width=0.48\linewidth]{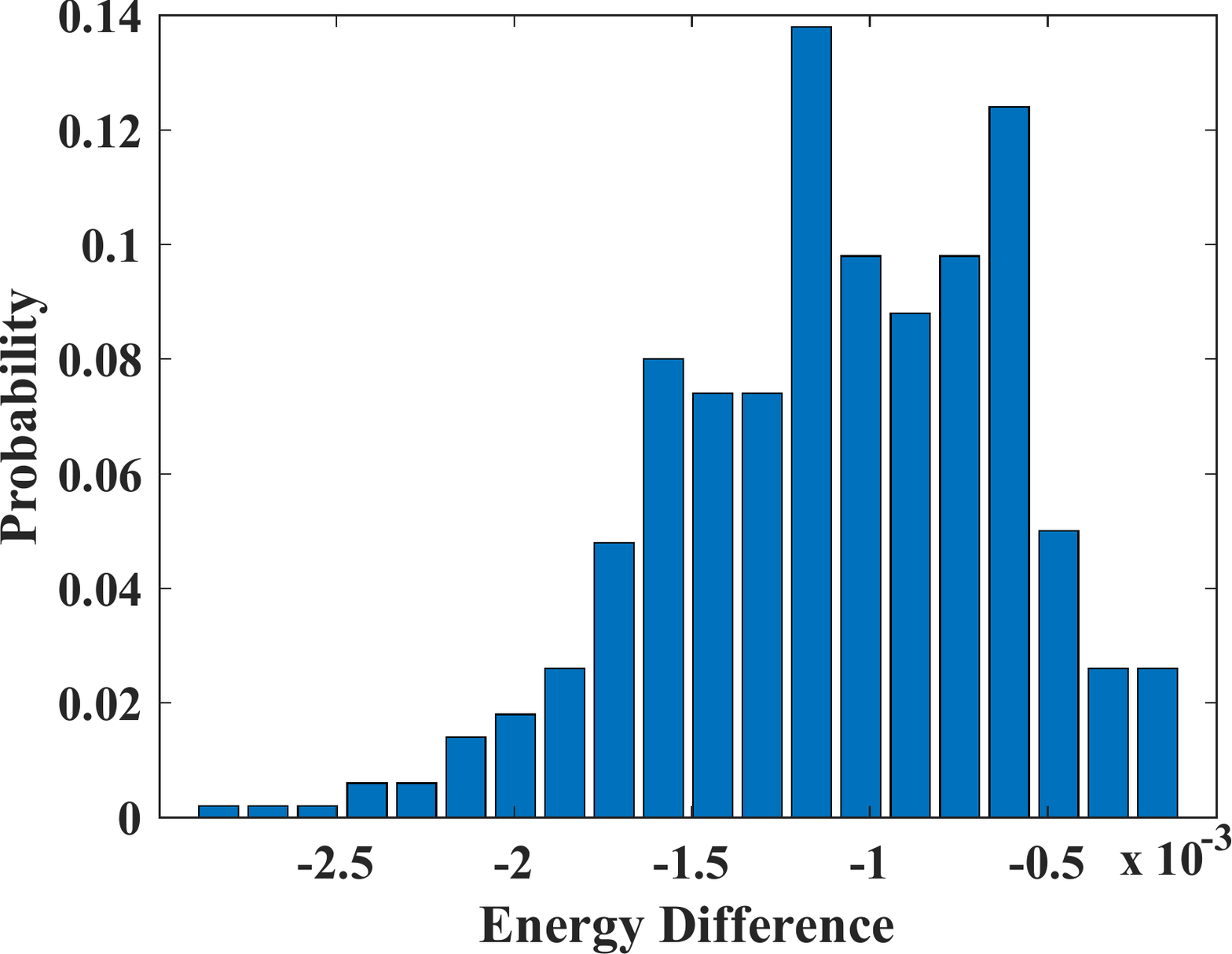}}
		\subfigure[$E_{RSnet} - E_{FS}$ distribution]{		
			\includegraphics[width=0.48\linewidth]{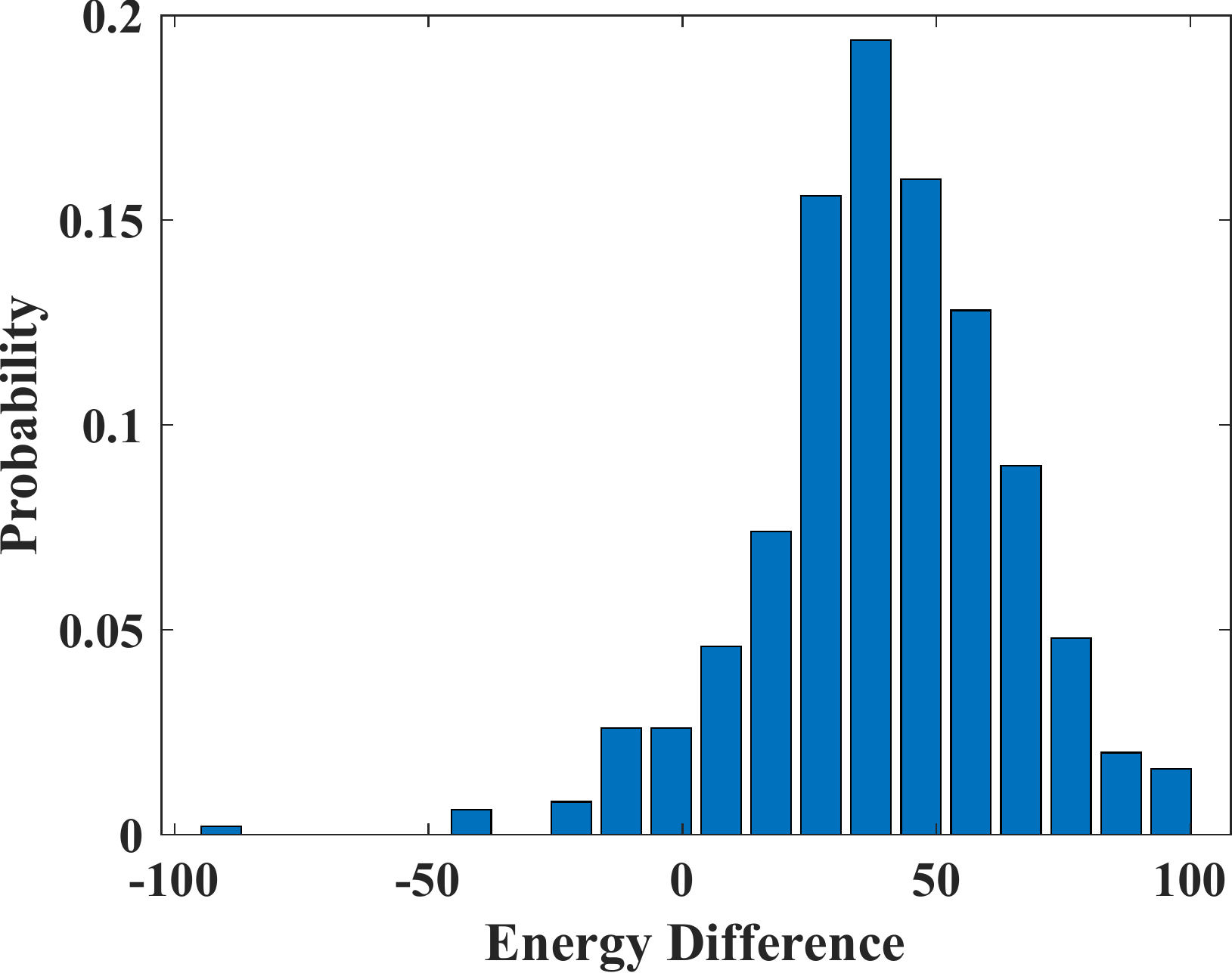}}		
		\subfigure[$PSNR_{RS} - PSNR_{FS}$]{		
			\includegraphics[width=0.48\linewidth]{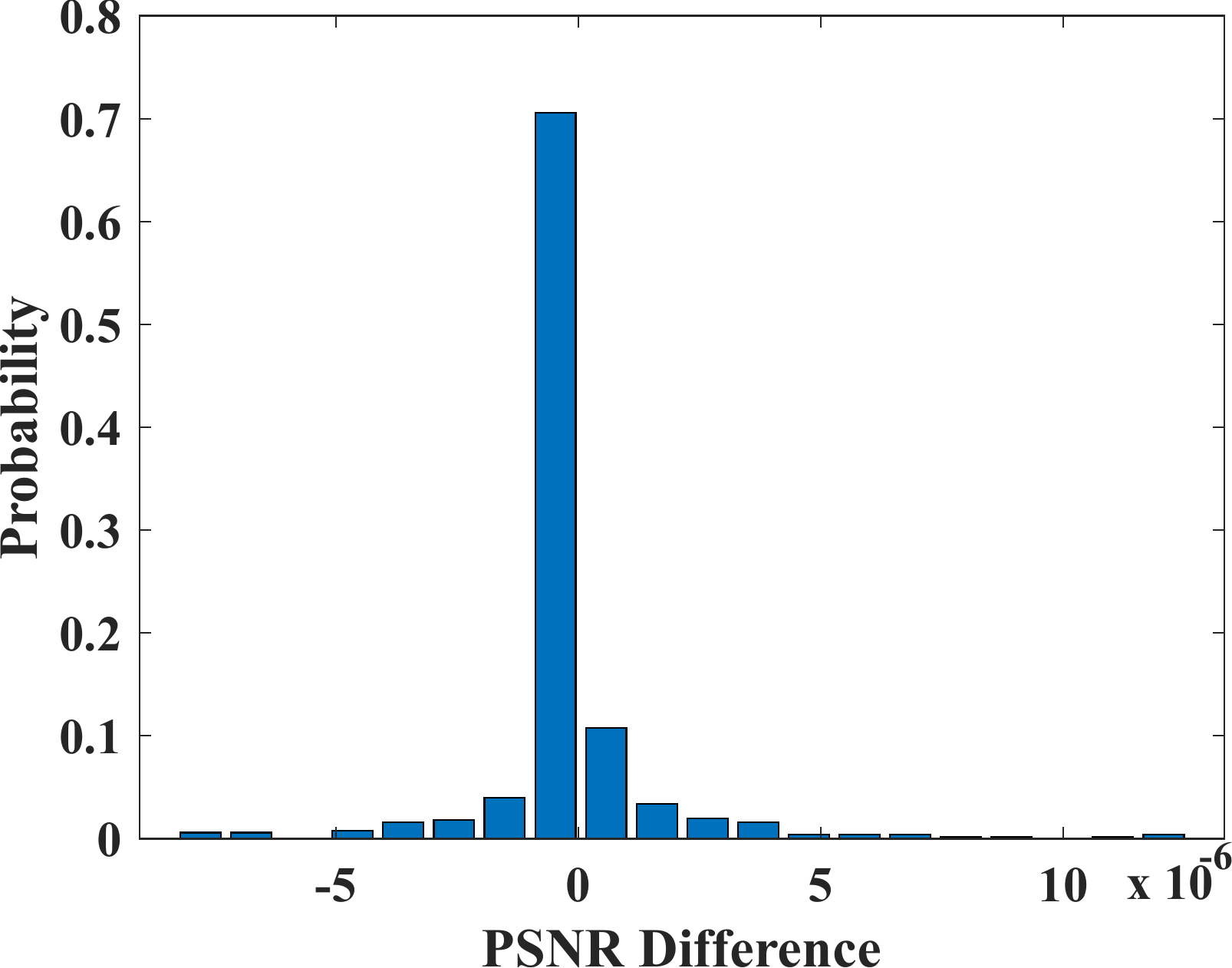}}
		\subfigure[$PSNR_{RSnet} - PSNR_{FS}$ ]{		
			\includegraphics[width=0.48\linewidth]{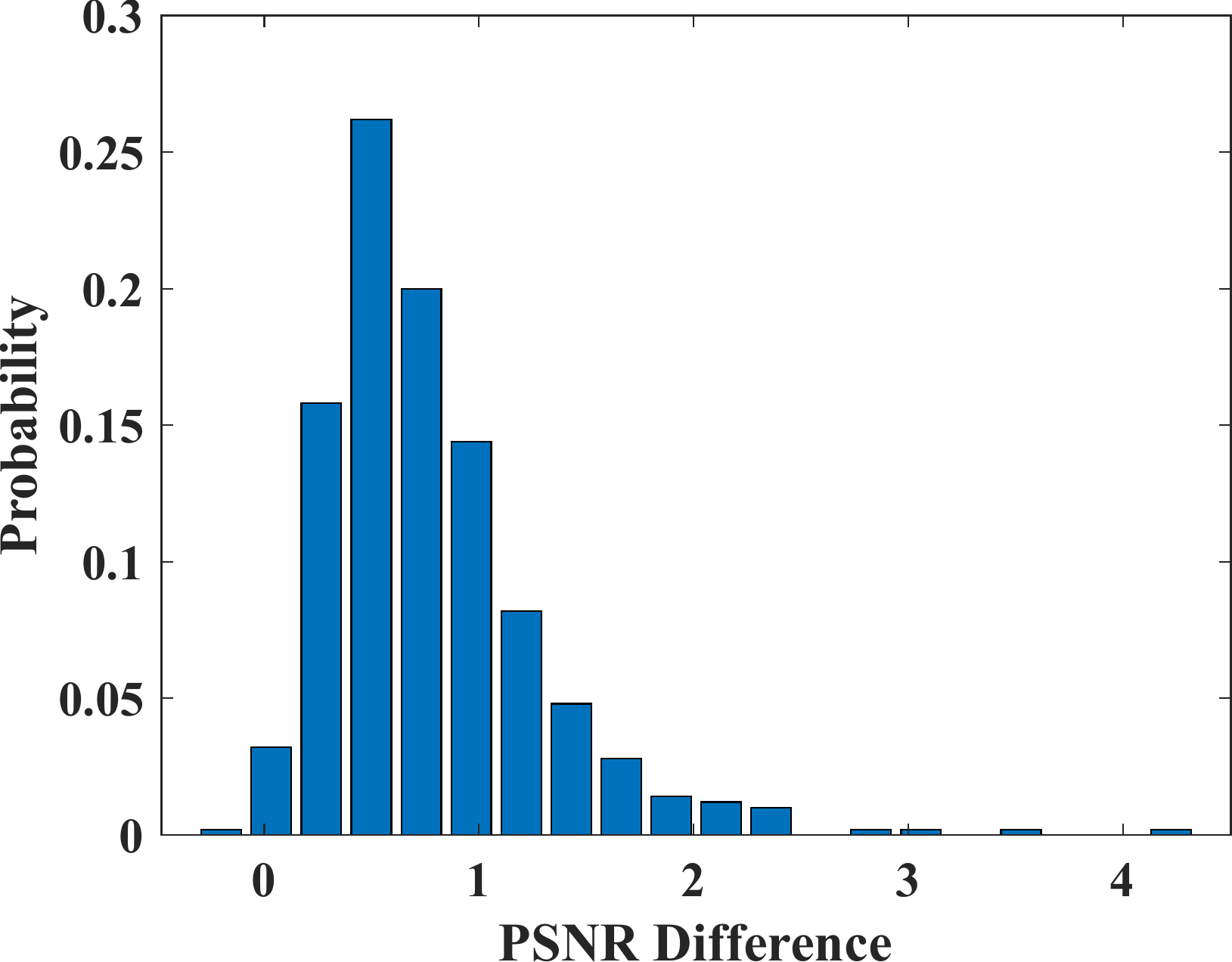}}
	\end{center}
	\caption{Compare FastSolver, RS and RSnet in image denoising task.}
	\label{fig:imagedenoisingperf}
\end{figure}

\subsection{Image denoising}
\label{sec:denoise}
In this experiment, we show the transferring ability of RSnet. We train our network on the noise free image patches with resolution $128\times 128$, but apply the trained network on noisy images with higher resolution. This indicates that the learned RSnet can be transferred for other tasks. We add noises with Gaussian distribution ($N(0,15^2)$) on the images from BSDS dataset. We apply FastSolver, RS and RSnet on these noisy images, respectively.

Two examples are shown in Fig.~\ref{fig:imagedenoising}. As can be seen in Fig.~\ref{fig:imagedenoising}, the results of FastSolver and RS are comparable but the result from RSnet is better. PSNR is calculated to show this quantitatively. There are two reasons for this. First, RSnet is an enhanced version of RS because RSnet has multi channels (16 in our case) while RS only has two channels (gradient operator). Second, RSnet was trained on the noise free images, which make the learned parameters produce similar results. 

We evaluate the denoising performance on 500 images of the dataset. In Fig.~\ref{fig:imagedenoisingperf}(a) and (b), we show the difference of the converged energy between RS, RSnet and FastSolver, respectively. We show the difference of their PSNR in Fig.~\ref{fig:imagedenoisingperf} (c) and (d), respectively . As can be seen, RS can converge to a smaller energy than FastSolver. Although RSnet gets higher energy, it can reach better PSNR. In this application, RSnet has the best performance due to the two reasons mentioned above.

\subsection{Ultrasonic image reconstruction}
In previous experiments, we have shown the performance of RS and RSnet on the problems where pixels are only affected by their local neighbors. In this section, we use our RS and RSnet to reconstruct ultrasonic images where each pixel has an influence on all other pixels (global behavior).

\begin{figure}[!htb]
	\begin{center}
		\subfigure[imaging]{\includegraphics[width=0.45\linewidth]{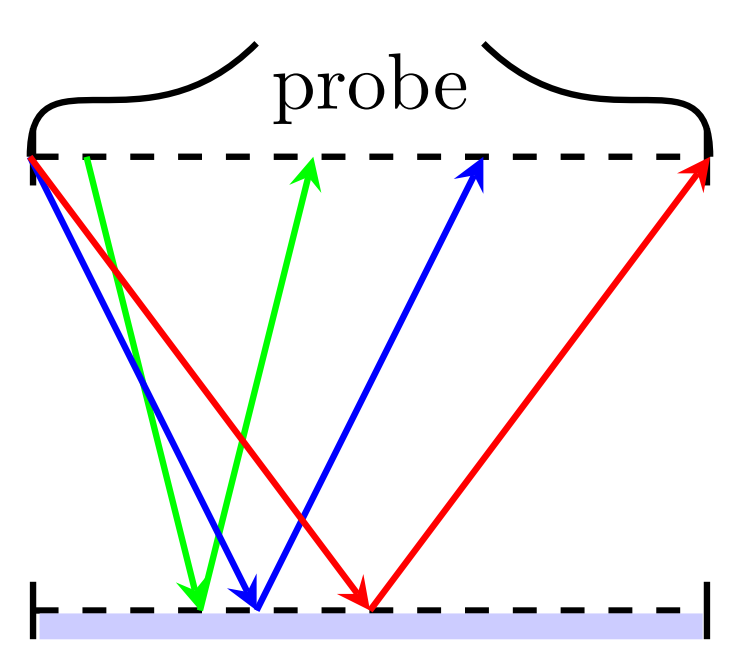}}~~~~~				
		\subfigure[discretization]{\includegraphics[width=0.4\linewidth]{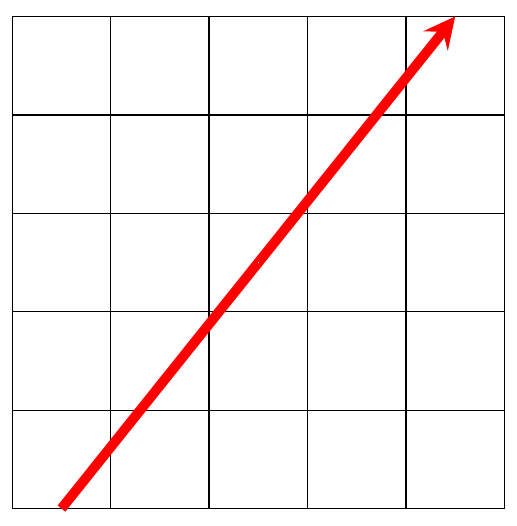}}	
	\end{center}
	\caption{Ultrasound imaging principle (a) and its discretization (b).}
	\label{fig:sos}
\end{figure}

The imaging principle of ultrasound is shown in Fig.~\ref{fig:sos} (a) and the imaging domain is discretized, shown in Fig.~\ref{fig:sos} (b). The average speed of sound in human tissue $S$ is 1540\,meters per second. We denote the contrast speed of sound image as $C(\vec{x})$ and then the speed of sound is $S+C(\vec{x})$. Letting $P(\vec{x})$ denote the sound wave traveling path, we can calculate its traveling time by
\begin{equation}
T=\int\limits_{\vec{x}\in P}\frac{\mathrm{d}P(\vec{x})}{S+C(\vec{x})}\,.
\end{equation} Letting $U=\frac{1}{S+C(\vec{x})}-\frac{1}{S}$ be the time delay compared with the average speed $S=1540$, we can get the total time delay
\begin{equation}
f=\int\limits_{\vec{x}\in P}(\frac{\mathrm{d}P(\vec{x})}{S+C(\vec{x})}-\frac{\mathrm{d}P(\vec{x})}{S})=\int\limits_{\vec{x}\in P}U\mathrm{d}P\,.
\end{equation} In practice, $\mathrm{d}P$ can be discretized on a mesh grid to form an imaging matrix $A$ (see Fig.~\ref{fig:sos} (b)). And the time delay $f$ can be measured by an ultrasound device. 

\begin{figure*}[!htb]
	\begin{center}
		\subfigure[Truth]{
			\includegraphics[width=0.177\linewidth]{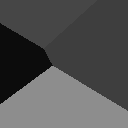}}
		\subfigure[Observation]{
			\includegraphics[width=0.177\linewidth]{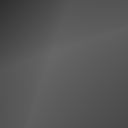}}
		\subfigure[{\scriptsize FS(38.97,0.99)}]
		{
			\begin{overpic}[width=0.177\linewidth]{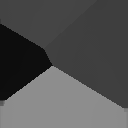}
				\put(10,43){\color{red}\vector(-1,-3){6}}
				\put(90,40){\color{red}\vector(1,-3){6}}
			\end{overpic}
		}
		\subfigure[{\scriptsize rs (38.97,0.99)}]{	
			\begin{overpic}[width=0.177\linewidth]{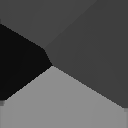}
				\put(10,43){\color{red}\vector(-1,-3){6}}
				\put(90,40){\color{red}\vector(1,-3){6}}
			\end{overpic}
		}
		\subfigure[{\scriptsize net(24.70,0.81)}]
		{
			\begin{overpic}[width=0.177\linewidth]{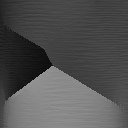}
				\put(10,43){\color{red}\vector(-1,-3){6}}
				\put(90,40){\color{red}\vector(1,-3){6}}
			\end{overpic}
		}
		%\subfigure[PSNR vs. $\sigma$]{		
		%	\includegraphics[width=0.21\linewidth]{images/ultrasonic_psnr_converge_1.pdf}}\\
		\subfigure[Truth]{
			\includegraphics[width=0.177\linewidth]{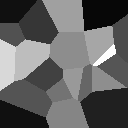}}
		\subfigure[Observation]{
			\includegraphics[width=0.177\linewidth]{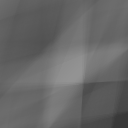}}
		\subfigure[{\scriptsize FS(38.42,0.99)}]
		{
			\begin{overpic}[width=0.177\linewidth]{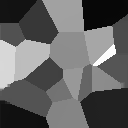}
				\put(8,52){\color{red}\vector(-1,-3){6}}
				\put(90,50){\color{red}\vector(1,-3){6}}
			\end{overpic}
		}
		\subfigure[{\scriptsize rs (38.42,0.99)}]{
			\begin{overpic}[width=0.177\linewidth]{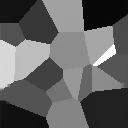}
				\put(8,52){\color{red}\vector(-1,-3){6}}
				\put(90,50){\color{red}\vector(1,-3){6}}
			\end{overpic}
		}	
		\subfigure[{\scriptsize net(18.28,0.80)}]
		{
			\begin{overpic}[width=0.177\linewidth]{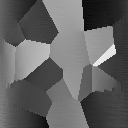}
				\put(8,52){\color{red}\vector(-1,-3){6}}
				\put(90,50){\color{red}\vector(1,-3){6}}
			\end{overpic}
		}	
	\end{center}
	\caption{Ultrasonic image reconstruction. From left to right: ground truth, the observed signal, result from FastSolver~\cite{jia2010fast}, result from RS and result from RSnet. The numbers in the brackets is PSNR (left) and SSIM (right), respectively.}
	\label{fig:imagerec}
\end{figure*}

To recover the $U$ (and $C$), we minimize following model
\begin{equation}
\label{eq:sos}
\frac{1}{2}\|AU-f\|_2+\lambda (w\|\nabla_x U\|_1+(1-w)\|\nabla_y U\|_1)\,,
\end{equation} where $0<w<1$ is a weight parameter since the uncertainty of each pixel along $x$ and $y$ axes are different~\cite{Sanabria2016}. We set $w=0.9$ according to this work.

We use FastSolver, RS and RSnet to solve this problem, respectively. We generate 500 simulated images as ground truth. To generate one image, we first generate a random region number $n$ and then generate $n$ random seeds in 2D. We use Voronoi diagram to obtain $n$ regions from these seeds. Then, we give each region a random number in $[0,255]$. For such image, according to the discretization in Fig.~\ref{fig:sos} (b), we obtain the corresponding observations. 

Our RSnet is trained on the same 10,000 natural image patches with $128\times128$ resolution (the same data set as described in previous sections), but the loss function is changed to Eq.~\ref{eq:sos}. The learning rate, batch size, epoch, etc. are the same as described in previous sections. This also shows the transfer property of RSnet since the training data set is natural images but the testing data set is piecewise constant biomedical images.

Two examples are shown in Fig.~\ref{fig:imagerec}, including the ground truth, observation and the reconstruction results from FastSolver, RS and RSnet, respectively. The reconstructed results from FastSolver and RS are almost the same since they reach the same global optimal solution. To evaluate the quality of the results, PSNR and SSIM are calculated and the values are shown in Fig.~\ref{fig:imagerec}. 

The reconstruction results have some issue at left and right boundaries, which is indicated by red arrows. The reason is that the ultrasound device does not have enough samples in the left and right boarder regions. This issue is not caused by solving algorithms but is the limitation from hardware. In practice, such boundary regions should be removed from the reconstructed images and the central region is the main concern. 

In this application, RSnet can achieve real time performance, much faster than the other two iterative algorithms because of the small number of blocks. Therefore, RSnet can be used in real-time imaging in our future work.  

%We also evaluate the reconstruction performance on 500 simulated images. In Fig.~\ref{fig:imagerecperf} (a) and (b), we show the difference of the converged energy and PSNR between FastSolver and RS. The energy different is on the level of $10^{-5}$, which almost reaches the single precision accuracy. There is no visual difference between results from these tow algorithms. This fact is also confirmed in Fig.~\ref{fig:imagerec}. 

%We add noises in the observations and by changing $\lambda$ from 1 to 20, we pick up the denoising results which has the maximum PSNR. The curves in Fig.~\ref{fig:imagedenoise} show the changing of PSNR when $\sigma$ is increased from 5 to 40. This confirms that the results of FastSolver and RS are comparable. 

\section{Conclusion}
We have developed a new algorithm for solving the total variation models, called residual solver. It works in gradient domain and implicitly optimizes the model. This new algorithm runs faster than the well-known fast solver~\cite{jia2010fast} and can reach the same global optimal solution. Its efficiency and effectiveness has been numerically confirmed by several experiments, including image smoothing, denoising and biomedical image reconstruction. 

Moreover, we unfold the new algorithm into a neural network. Our network is unsupervised and thus can be trained without knowing the ground truth. Our network does not contain any batch normalization or drop out layers. Therefore, it can be trained on low resolution images but applied on high resolution images. Our numerical experiments have confirmed this property. The results from our network can be compared with the counterpart from iterative algorithms because they use the same loss function.  

A small number of blocks in our neural network can reach the global optimal solutions. This property is different from iterative algorithms, such as FastSolver and RS, which require a large number of iterations to converge. Thanks to the high performance, our network is suitable for many real-time applications in practice. Such high performance might lead to real-time high resolution video processing algorithms. 

We believe that the proposed RS and RSnet can be applied on a large range of image processing problems for several reasons. First, the loss function is well established in the field and has been shown successful in the past decades. Therefore, the proposed solvers can be applied on a large range of applications. Second, our RS and RSnet can reach the global optimal solutions, as confirmed in this paper. RS is suitable for tasks with high accuracy requirement. RSnet is suitable for real-time applications. 

In future work, we will investigate their applications in the tasks where total variation is used as regularization, for example, image deblurring~\cite{Gong2018}, inpainting~\cite{Yin_2019_CVPR}, image smoothing~\cite{GONG2019329}, super resolution, image curvature optimization~\cite{gong:cf}, dehazing~\cite{YIN2019315}, image fusion, motion estimation~\cite{Gong2019}, computed tomography in X-ray imaging~\cite{8803492}, etc.

\section*{Acknowledgment}
This work was supported in part by the National Natural Science Foundation of China under Grant 61907031.

% or




%\bibliographystyle{IEEEtran}
\bibliographystyle{IEEEtran}
\bibliography{IEEEabrv,IP}



%\vfill


\end{document}